\PassOptionsToPackage{table}{xcolor}
\documentclass[runningheads]{llncs}

\usepackage{eccv}

\usepackage{eccvabbrv}

\usepackage{graphicx}
\usepackage{booktabs}

\usepackage[accsupp]{axessibility}  
\usepackage{hyperref}
\usepackage{amsmath}
\usepackage{url} 
\usepackage{graphicx}
\usepackage{adjustbox}
\usepackage{wrapfig}
\usepackage{floatrow}
\usepackage{subcaption}
\usepackage{listings}
\usepackage{amssymb}
\usepackage{algorithmicx}
\usepackage{enumitem}
\usepackage{caption}
\usepackage{algorithm}
\usepackage{algpseudocode}
\usepackage[toc,page,header]{appendix}
\usepackage{bytefield}
\usepackage{textcomp}
\usepackage{xcolor}
\usepackage{booktabs}
\usepackage{tcolorbox}
\usepackage{multirow}
\usepackage[autostyle=true, style=english]{csquotes}
\MakeOuterQuote{"}

\usepackage{amsmath,amsfonts,bm}

\def\eqref#1{equation~\ref{#1}}

\def\1{\bm{1}}

\DeclareMathAlphabet{\mathsfit}{\encodingdefault}{\sfdefault}{m}{sl}
\SetMathAlphabet{\mathsfit}{bold}{\encodingdefault}{\sfdefault}{bx}{n}

\definecolor{gaincolor}{HTML}{e9fff6}
\definecolor{losscolor}{HTML}{ffd9e6}
\newcommand{\gain}[1]{\cellcolor{gaincolor}{#1}}

\definecolor{hkured}{HTML}{E8304C}
\definecolor{myblue}{RGB}{210, 225, 255}
\definecolor{lightred}{RGB}{255, 182, 193}   
\definecolor{lightgreen}{RGB}{144, 238, 144}  
\definecolor{lightblue}{RGB}{173, 216, 230} 
\definecolor{lightergray}{gray}{0.90}

\usepackage{hyperref}

\usepackage{orcidlink}

\begin{document}

\title{SRUM: Fine-Grained Self-Rewarding for Unified Multimodal Models}

\titlerunning{SRUM}

\author{
Weiyang Jin\inst{1,2,5}\textsuperscript{*} \and
Yuwei Niu\inst{3}\textsuperscript{*} \and
Jiaqi Liao\inst{2} \and
Chengqi Duan\inst{1,2} \and
Aoxue Li\inst{4} \and
Shenghua Gao\inst{2} \and
Xihui Liu\inst{1,2,5}\textsuperscript{$\dagger$}}

\authorrunning{W.~Jin et al.}

\institute{
HKU MMLab \and
The University of Hong Kong \and
Peking University \and
Noah's Ark Lab, Huawei \and
Shenzhen Loop Area Institute\\
\email{weiyangjin@connect.hku.hk}}

\maketitle
\begingroup
\renewcommand{\thefootnote}{}
\footnotetext{\textsuperscript{*}Equal contribution. \textsuperscript{$\dagger$}Corresponding author: xihuiliu@eee.hku.hk. Project page: \url{https://waynejin0918.github.io/srum_web/}.}
\endgroup

\begin{abstract}
Recently, remarkable progress has been made in Unified Multimodal Models (UMMs), which integrate vision-language generation and understanding capabilities within a single framework. However, a model's strong visual understanding often fails to transfer to visual generation: it may correctly judge prompt-image alignment while failing to generate a faithful image from the same prompt. This raises a compelling question: Can a model improve itself by using its understanding module to reward its generation module? We introduce SRUM, a self-rewarding post-training framework directly applicable to existing UMMs of various designs. SRUM creates a feedback loop where the model's own understanding module acts as an internal ``evaluator'', providing corrective signals to improve generation without additional human-labeled data or external reward models. To provide comprehensive feedback, SRUM uses a global-local dual reward system: a \textbf{global reward} ensures overall visual semantics and layout, while a \textbf{local reward} refines fine-grained, object-level fidelity. SRUM shows strong generalization, boosting performance on T2I-CompBench from 82.18 to \textbf{88.37} and on T2I-ReasonBench from 43.82 to \textbf{46.75}. Overall, our work establishes a powerful paradigm for enabling a UMM's understanding module to guide and enhance its own generation via self-rewarding. 
\end{abstract}

\section{Introduction}
\label{sec:intro}
Text-to-Image (T2I) models have achieved remarkable progress in generating high-quality and diverse images from given prompts~\cite{dalle1, imagen, sdxl}. However, they often fail to accurately interpret instructions involving world knowledge, complex spatial relationships, detailed attribute binding, or compositional reasoning~\cite{huang2023t2i}. These limitations point to a fundamental lack of deep semantic understanding in standard T2I models. To address this challenge, researchers have developed Unified Multimodal Models (UMMs), which integrate both understanding and generation capabilities within a single framework~\cite{vilau2024, janus2024, dongdreamllm, show-o}. By sharing a common backbone, UMMs possess the inherent potential for synergy, offering a promising path to resolve the comprehension challenges that plague traditional T2I models.

\begin{figure}[htbp]
\centering
\makebox[\textwidth][c]{\includegraphics[width=0.92\textwidth]{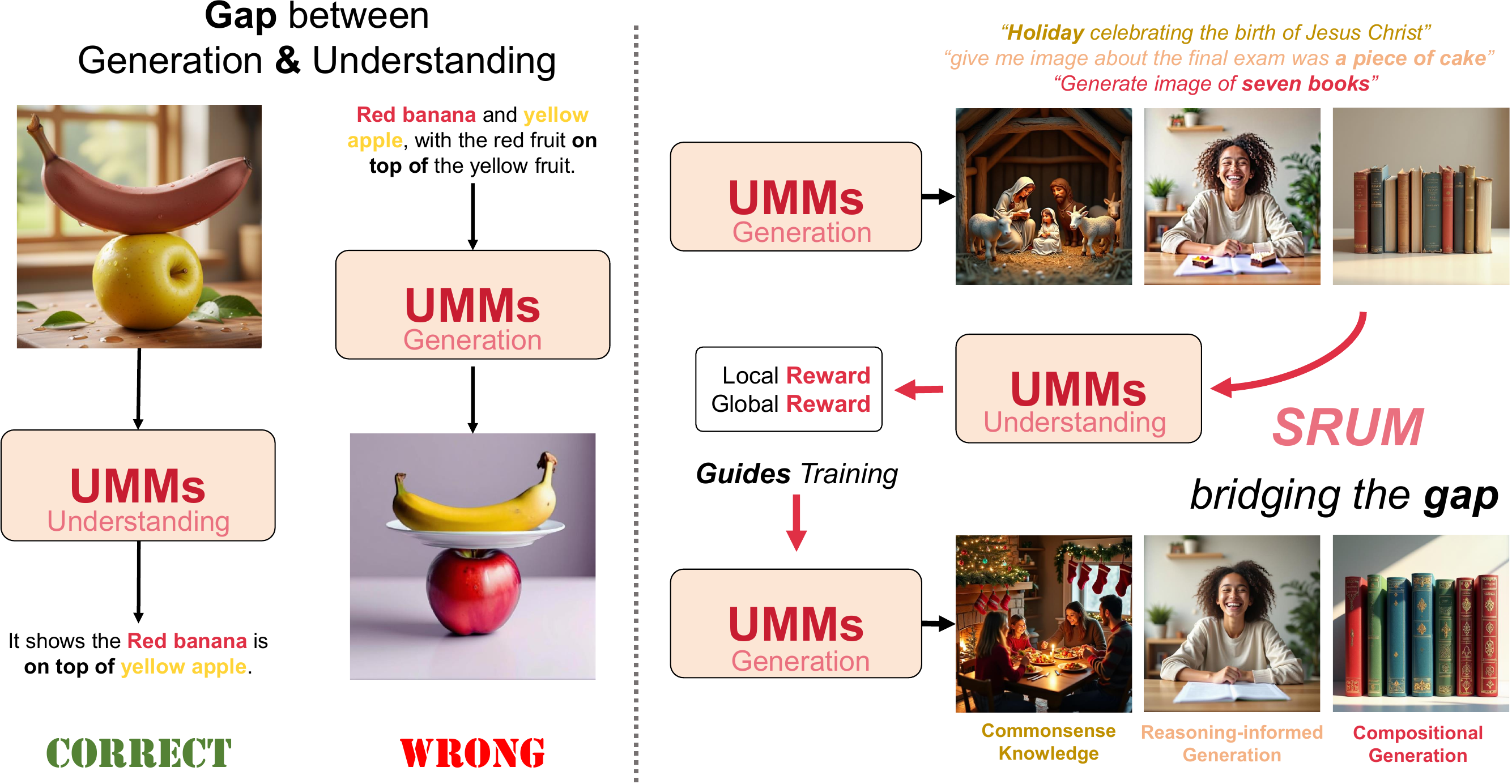}}
\caption{The example on the left suggests that the current UMMs' understanding module has exceeded the capability of its generation module: the generation module is prone to producing incorrect candidate images based on a given prompt in relevant scenarios, a situation which the understanding module can reasonably identify. This not only highlights a gap between understanding and generation but also reveals the potential for understanding to guide generation. Inspired by this insight, we propose \textbf{SRUM} to bridge this gap, particularly in complex generation domains.}
\label{fig:teaser}
\end{figure}

Despite their advanced architecture, a fundamental paradox plagues current UMMs: their capacity to generate falls far behind their ability to understand~\cite{tong2024metamorph,januspro2025,pan2025transfer,xie2025show,emu3}. For instance, a model can often correctly judge the alignment between a detailed prompt and a complex image, yet be incapable of generating a faithful image from that same prompt (\Cref{fig:teaser}). This persistent gap between understanding and generation suggests that the key to unlocking better generation lies within the model itself.

To address this challenge, we propose bridging this module gap through self-rewarding. We introduce \textbf{S}elf-\textbf{R}ewarding for \textbf{U}nified Multimodal \textbf{M}odels (SRUM), a novel post-training framework designed to create a synergistic feedback loop within the model itself. Our core insight is that the solution lies within the UMMs' own architecture. By treating the generation module as a ``student'' and the more capable understanding module as an internal ``teacher'' or ``evaluator,'' we establish a practical self-rewarding system for improvement, without external reward models, human labels, or additional image data during the training phase.

Furthermore, to effectively guide the generation of complex scenes, a reward signal should provide multi-scale feedback. As our ablation studies confirm, a single, holistic score is insufficient because it fails to provide the fine-grained corrective signals needed for detailed improvement. Therefore, we propose a global-local dual reward framework. The global reward evaluates high-level compositional coherence to ensure overall scene plausibility. Concurrently, the local reward targets object-level details, optimizing attribute binding and spatial arrangements. This synergistic design enables SRUM to enhance the performance of the base model on complex generation tasks.

Through extensive experiments, we demonstrate that our approach significantly improves the composition, reasoning, and visual fidelity of UMMs, showing strong generalization across in-domain and out-of-domain settings. SRUM achieves SOTA results on T2I-CompBench and T2I-ReasonBench, improving the overall score of a strong baseline model from 82.18 to 88.37 in composition and from 43.82 to 46.75 in reasoning. Our key contributions can be summarized as follows:

\begin{itemize}[leftmargin=2em]
\item We are the first to propose a comprehensive self-rewarding framework for UMMs at the post-training stage, successfully bridging the gap between their understanding and generation.
\item We introduce a novel dual reward design that combines global compositional assessment with local object-level feedback, providing solid and multi-scale guidance during model training.
\item We achieve better performance on complex compositional generation and demonstrate strong generalization. Ultimately, SRUM establishes a powerful paradigm for a UMM's understanding module to guide its own generation module toward self-improvement.
\end{itemize}
\section{Related Works}

\subsection{Architectures for Unified Multimodal Models}
Unified Multimodal Models (UMMs) have emerged as a prominent research direction, aiming to integrate diverse tasks like visual understanding and generation within a single, end-to-end trained architecture. This consolidation seeks to foster synergy across modalities and reduce systemic complexity. Recent architectural paradigms can be broadly categorized. The \textbf{Purely Autoregressive (AR)} approach extends the next-token prediction paradigm of LLMs to visual data, treating images as a sequence of discrete tokens~\cite{chameleon, emu3}. A key refinement in this area involves decoupling the visual encoders, using a semantic encoder for understanding tasks while retaining a reconstruction-based tokenizer for generation, as demonstrated by Janus~\cite{janus2024}. Show-O further refines this by integrating a discrete-diffusion schedule to improve token prediction~\cite{show-o}. More prevalent are hybrid architectures that combine the strengths of AR and diffusion models. One major category consists of \textbf{Sequential AR-Diffusion} models, where an AR component generates an intermediate representation that conditions a diffusion-based decoder. In some variants, a pre-trained MLLM is kept frozen for reasoning, and its features are routed via learnable queries or hidden states to an external image generator~\cite{tong2024metamorph, shi2024llamafusion, lin2025uniworld}. This cascaded design effectively leverages powerful existing models. A more integrated approach uses a \textbf{Unified Transformer Backbone}~\cite{monoformer2024,chen2024diffusion}, where both AR and diffusion objectives are optimized simultaneously within a single transformer. To improve scalability, the \textbf{Mixture-of-Transformers (MoT)} paradigm has been introduced~\cite{mot, deng2025emerging}. This approach, exemplified by Bagel, employs a sparse, modular design where specialized experts handle different modalities but share information through a common attention mechanism. Our work, SRUM, inherits this flexible MoT framework, demonstrating the versatility of our method on advanced UMMs.

\subsection{Post-Training Stage in UMMs}
In addition to architectural innovations, considerable research has focused on the post-training stage to enhance the generative abilities of UMMs. Methods such as Chain-of-Thought (CoT) and test-time verification introduce explicit reasoning steps or iterative output validation~\cite{guo2025can,fang2025got,duan2025got}. However, these often depend on external models and do not fundamentally improve the native generative capacity of UMMs. Reinforcement learning techniques, including Direct Preference Optimization (DPO), leverage human or automated feedback to refine generation policies. While effective, these require carefully curated paired data and delicate advantage-function tuning with text-dependent rewards~\cite{rafailov2023direct,guo2025r1,qu2025silmm}. Reconstruction Alignment (RecA) introduces a post-training method based on reconstruction loss, yielding improved semantic understanding~\cite{xie2025reconstructionalignmentimprovesunified}. Some work has also attempted to use rule-level rewards for guidance, but such rewards are not universal and must be redesigned for different tasks~\cite{hong2025reinforcing,mao2025unirl,han2025self}. In contrast, SRUM leverages the model's inherent understanding to score self-generated samples, thereby enhancing performance without external reward models or human preference labels.

\subsection{Self-Rewarding in Understanding Models}
Self-rewarding mechanisms have emerged as a significant paradigm for enhancing the understanding and reasoning capabilities of MLLMs. These approaches aim to reduce reliance on external preference data by enabling models to generate their own reward signals, thereby facilitating continuous self-improvement. For instance, CSR~\cite{zhou2024calibrated} achieves zero-cost self-enhancement through iterative online DPO with visual constraint rewards. SRPO~\cite{choi2024self} introduces a two-stage reflective reward mechanism, significantly improving the quality of reflection and answer accuracy in complex reasoning tasks. R1-Reward leverages process consistency rewards and stable reinforcement learning algorithms to enhance long-range reasoning stability~\cite{guo2025r1}. Collectively, these works signal a paradigm shift from external rewards to self-criticism and optimization. However, they tend to focus on a single dimension of feedback. Our SRUM framework proposes a more holistic approach. It distinguishes itself by incorporating a global-local dual reward system designed to provide a more comprehensive training signal. 
\section{SRUM: Self-Rewarding for Unified Multimodal Models}
To drive self-improvement where the model's understanding capabilities guide its generation abilities, we establish a multi-stage self-rewarding process. First, the Unified Multimodal Models (UMMs) generate high-quality candidate images with corresponding bounding boxes (as detailed in Section \ref{sec:data_gen}). Next, these candidates are evaluated by a global-local judgment framework that assesses both overall composition and fine-grained details, producing a holistic reward signal (Section \ref{sec:reward}). Finally, the cached rewards directly inform a reward-weighted training process, which enables targeted, region-specific optimization and reduces reward hacking (Section \ref{sec:weighted}).
\begin{figure}[H]      
   \vspace{-2em}
  \hspace*{0.01\textwidth}
  \includegraphics[width=0.88\textwidth]{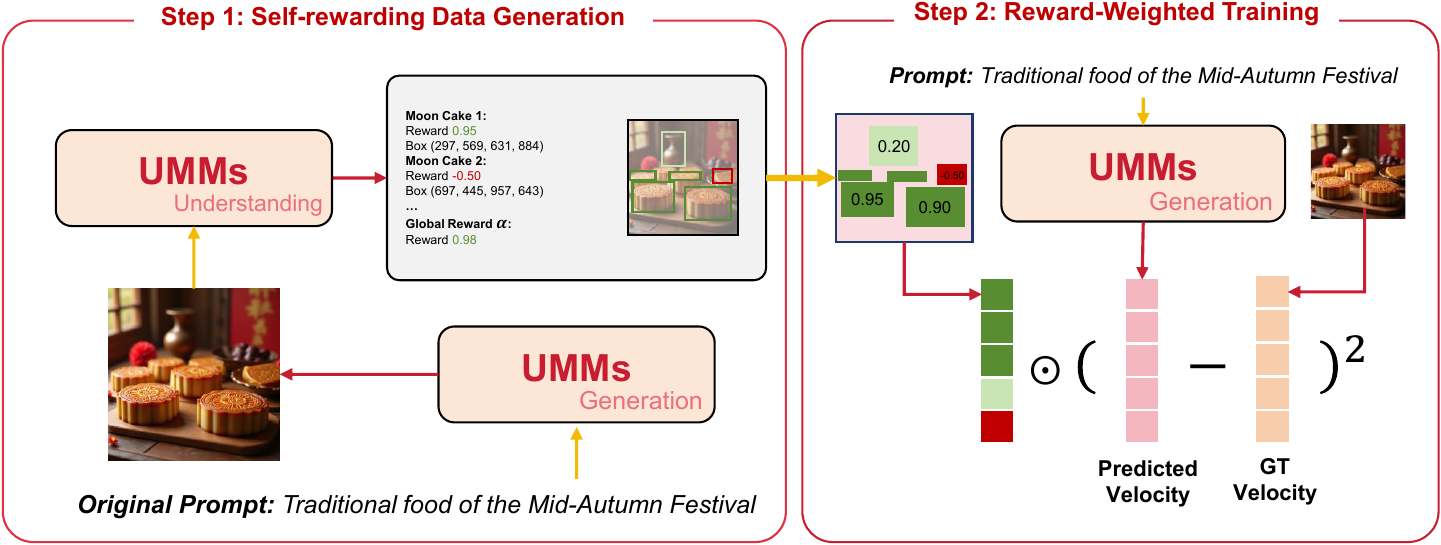}
  
  \caption{Showcase of the SRUM pipeline. It consists of two main stages: Self-Rewarding Data Generation and Reward-Weighted Training. The first stage generates high-quality data and scores it to produce a reward for the next training stage for self-improvement.}
  \label{fig:srum_ppl}
\end{figure}
\vspace{-2em}
\subsection{Image Candidates and Bounding Box Generation}

\label{sec:data_gen}
We developed a self-data generation pipeline that enables our model to create its own training data, removing the need for external image resources. It starts with the UMM using its ``think'' mode (a form of CoT) to generate semantically rich images~\cite{deng2025emerging,wang2025mint}. For Bagel, an external segmentation model proposes spatial supports for grounding~\cite{kirillov2023segment}; for BLIP3o, native grounding can be used instead. In both cases, the UMM verifies boxes, matches prompt-relevant regions, and assigns all semantic rewards, so the external component is only a localization aid rather than an external reward model.

\subsection{Rewarding Process}

\label{sec:reward}
\textbf{Self-Judgment for Reliable Rewarding}. A cornerstone of self-improvement is enabling the model's internal understanding module to serve as a stable and reliable ``evaluator''. To ensure the scores it generates are consistently trustworthy, we designed a comprehensive self-judgment mechanism to assess image quality and prompt alignment~\cite{xu2023imagereward, zhang2023blind, lin2024evaluating, ghosh2023geneval}. This dual-level judgment is key to guaranteeing a thorough assessment. First, a local judgment evaluates object-level fidelity and artifacts on a strict $[-1.0, 1.0]$ scoring scale. A mandatory ``Reason'' field elicits an interpretable rationale for the score, akin to chain-of-thought prompting~\cite{guo2025can, fang2025got}, which further bolsters the reliability of the process. We enforce semantic grounding by verifying that identified objects correspond to prompt keywords, and a non-linear penalty maps severe distortions to a high-penalty negative range (e.g., -0.9 to -0.5) to better reflect human visual sensitivity. Subsequently, a global judgment evaluates the holistic composition and spatial alignment with the prompt's intent. Crucially, for prompts lacking specific compositional directives (e.g., ``a picture of a tree''), a neutral score range (e.g., -0.4 to 0.4) is applied, ensuring a fair assessment.

\textbf{Reward Generation for Training}. To serve as the core learning signal for self-improvement, the self-judgment scores are converted into a dense reward map. This step leverages the UMM's grounding capabilities to produce fine-grained local rewards for prompt-relevant regions and a single global reward for the whole image. The global score is normalized to $[0,1]$ to avoid spurious positive signals from multiplying two negative values (Appendix~\Cref{app:data_curation}). During reward generation, the understanding module is frozen and receives no gradients. SRUM is therefore offline: it generates candidates, caches rewards, and then updates the trainable generation/flow parameters.

\subsection{Reward-Weighted Training}
\label{sec:weighted}
The reward-weighted training stage is where the model achieves self-improvement through training with rewards. The core objective is to translate the capabilities of the understanding module directly into the functionality of the generation module. By using fine-grained local rewards and layout-aware global rewards to weight the training objective, we guide the generator to learn more detailed and accurate patterns from the original data. This process is the key to bridging the gap between the model's understanding and generation components, enabling the generator to benefit from the insights of the evaluator. The mechanism for this goal is a reward-weighted training objective, centered on the loss term $\mathcal{L}_{\text{r}}$. This term operates on the model's velocity prediction $v_\theta$, a standard practice in flow-based frameworks~\cite{liuflow, lipman2022flow}. The loss is modulated by two cached feedback signals from the frozen understanding module: a regional reward map $R \in [-1, 1]$ for localized refinement and a global scalar $\alpha$ for overall compositional quality. The product of these signals, $\alpha \cdot R$, weights the squared error between the predicted velocity $v_\theta$ and the target velocity derived from the original latent $x_0^{\text{gt}}$. This allows for fine-grained control, encouraging preservation where feedback is positive ($\alpha \cdot R > 0$) and promoting change where it is negative ($\alpha \cdot R < 0$):
\begin{equation}
    \mathcal{L}_{\text{r}} = \mathbb{E} \left[ \alpha \cdot R \odot \left( v_\theta - (\epsilon - x_0^{\text{gt}}) \right)^2 \right]
    \label{eq:l_reward}
\end{equation}

Second, to ensure the model's output conforms to the desired overall structure and to prevent reward hacking, we introduce a reference constraint term, $\mathcal{L}_{\text{ref}}$. This term acts as a regularizer, penalizing the squared $\ell_2$ distance to the target velocity of the artifact-free latent $x_0^{\text{gt}}$:
\begin{equation}
    \mathcal{L}_{\text{ref}} = \mathbb{E} \left[ \left\| v_\theta - (\epsilon - x_0^{\text{gt}}) \right\|^2 \right]
    \label{eq:l_cons}
\end{equation}

The final training objective is a weighted sum of these two losses, balanced by a hyperparameter $\lambda_{\text{c}}$. This is the complete objective used by SRUM; no hidden online reward loss or evaluator update is applied. The composite design enables targeted local refinement while maintaining global coherence:
\begin{equation}
\label{eq:weighted_objective_final}
    \mathcal{L}_{\text{Total}} = \mathcal{L}_{\text{r}} + \lambda_{\text{c}} \cdot \mathcal{L}_{\text{ref}}
\end{equation}
\section{Analysis of Self-Rewarding}
We validate our Self-Rewarding for Unified Multimodal Models (SRUM) method across various unified multimodal models (UMMs) and evaluation benchmarks. In particular, we investigate the following aspects:
     
\begin{itemize}[leftmargin=1em]
    \item \textbf{Generality and Performance:} SRUM achieves better performance on composition and delivers consistent gains across frameworks. (\Cref{tab:compbench_final})
    \item \textbf{Component Efficacy:} Ablation studies confirm each component of the SRUM framework makes a critical contribution to the overall performance. (\Cref{fig:combined_ab_kl})
    \item \textbf{Generalization:} SRUM demonstrates in-domain and out-of-domain generalization, indicating that improvements in generation can be transferred from understanding. (\Cref{tab:ablation_attributes}, \Cref{fig:wise}, \Cref{tab:reason})
\end{itemize}
     
\subsection{Experimental Setup}
     
\label{subsec:exp_setup}
\textbf{Model Architectures.} We evaluate SRUM on two powerful open-source UMMs. All experiments are conducted as a post-training phase, starting from the official pre-trained weights. Bagel~\cite{deng2025emerging} is a versatile UMM that serves as our primary model for comprehensive analysis, including main results, ablation studies, and generalization tests. We evaluate both its standard and CoT inference modes. BLIP3o~\cite{chen2025blip3,wang2023detecting,radford2021learning} is another current SOTA UMM used to validate the generality and effectiveness of SRUM with frozen MLLM training. We focus on these two families because some AR-type models, such as Show-O and Janus, can exhibit biases stemming from under-leveraged generation or understanding capabilities.

\textbf{Datasets and Benchmarks.} Our experiments leverage several specialized datasets for training and evaluation to ensure a thorough and multi-faceted analysis. For consistent and objective scoring across all generation benchmarks, we employ QwenVL-2.5-32B/QwenVL-2.5-72B~\cite{qwen2.5-vl} as the designated multimodal evaluator. It is crucial to clarify that these external models serve strictly as objective evaluation metrics (akin to accuracy in classification or mIoU in segmentation) and are completely excluded from our core self-rewarding training pipeline. We also report traditional metrics following prior work~\cite{chen2025blip3}. Our experiment begins with instruction data sourced from the T2I-CompBench training set~\cite{huang2023t2i}. For our primary evaluation, we use the standard split of the same benchmark to compare SRUM-enhanced models against leading T2I and UMM baselines. To assess generalization, we evaluate the model's in-domain transferability on GenEval~\cite{ghosh2023geneval}, which includes similar compositional challenges, and WISE~\cite{niu2025wise}, which evaluates knowledge-informed generation. Furthermore, we evaluate broader out-of-domain reasoning-informed capabilities on T2I-ReasonBench~\cite{sun2025t2i}, a benchmark containing complex prompts that require knowledge beyond the training distribution.
     
\subsection{Main Results}
\begin{table*}[ht]
    \centering
    \caption{\textbf{Comprehensive T2I-CompBench Results.} This table includes T2I and UMMs models. Models incorporating the SRUM are denoted with \textbf{+SRUM}. \textbf{Bold} indicate the highest score in each column. \gain{Green} values indicate the improvements.}
    \label{tab:compbench_final}
    \vspace{-0.5em}
    \resizebox{\textwidth}{!}{
        \setlength{\tabcolsep}{3pt} 
        \renewcommand{\arraystretch}{1.2}
        \begin{tabular}{lccccccccc}
            \toprule
            \textbf{Model} & \textbf{3D Spat.} & \textbf{Color} & \textbf{Compl.} & \textbf{Nonspat.} & \textbf{Num.} & \textbf{Shape} & \textbf{Spatial} & \textbf{Text.} & \textbf{Overall} \\
            \midrule
            \multicolumn{10}{c}{\textit{T2I Models}} \\ \midrule
            FLUX.1-dev & 76.39 & 90.63 & 83.51 &\bfseries 87.47 & \bfseries75.30 & 80.20 & 84.23 & 87.07 & 83.10 \\
            FLUX.1-schnell & \bfseries 79.38 & 84.53 & 81.96 & 85.55 & 72.82 & 82.20 & 85.49 & 86.38 & 82.29 \\
            SD-3-medium & 77.83 & \bfseries91.63 & \bfseries84.73 & 86.12 & 72.80 & \bfseries83.72 & \bfseries 88.20 & \bfseries89.03 & \bfseries84.26 \\
            SD-xl-base-1 & 72.25 & 77.75 & 75.00 & 85.28 & 57.14 & 72.18 & 77.08 & 78.38 & 74.38 \\
            \midrule \multicolumn{10}{c}{\textit{Unified Multimodal Models}} \\ \midrule
            Janus-Pro & 76.17 & 84.25 & 80.28 & 80.47 & 56.43 & 65.14 & 79.67 & 69.67 & 74.01 \\
            Show-O2 & \bfseries 88.61 & 87.73 & 87.88 & 85.91 & 69.74 & 73.99 & 86.60 & 82.17 & 82.83 \\
            OmniGen2 & 82.21 & 92.22 & 86.87 & 88.51 & 72.00 & 83.95 & 90.07 & \bfseries 90.88 & 85.84 \\ \midrule
            BLIP3o & 81.73 & 89.92 & 85.55 & 84.78 & 71.67 & 83.75 & 92.47 & 87.45 & 84.66 \\
            \rowcolor{gray!15}
            \quad \textbf{+SRUM} & 83.78 & 90.22 & 86.57 & 85.10 & 74.52 & 85.44 & \bfseries 93.88 & 86.52 & \gain{85.75} \\
            \midrule
            Bagel & 77.98 & 89.30 & 83.32 & 85.03 & 70.40 & 81.94 & 81.52 & 87.93 & 82.18 \\
            \rowcolor{gray!15}
            \quad \textbf{+SRUM} & 83.10 & 92.90 & 88.69 & 88.47 & 78.52 & 84.23 & 86.92 & 89.57 & \gain{86.55} \\
            \midrule
            Bagel$_{\text{(CoT)}}$ & 84.66 & 88.85 & 86.10 & 85.64 & 75.36 & 84.33 & 82.71 & 88.07 & 84.46 \\
            \rowcolor{gray!15}
            \quad \textbf{+SRUM} & 88.60 & \bfseries 92.90 & \bfseries 91.31 & \bfseries 90.48 & \bfseries 80.12 & 84.47 & 89.93 & 89.15 & \gain{\bfseries 88.37} \\
            \bottomrule
        \end{tabular}
    }
\end{table*}
   \vspace{-1em}

\begin{table*}[htbp]
    \centering
    \caption{\textbf{Detailed T2I-CompBench Evaluation via Traditional Metrics.} This table compares various UMMs models using specialized traditional evaluators. Models incorporating the SRUM are denoted with \textbf{+SRUM}. \textbf{Bold} indicates the highest score in each column. All values are reported in percentage (\%) (higher is better $\uparrow$).}
    \label{tab:appendix_traditional_metrics_simplified}
    \vspace{-0.5em}
    \resizebox{\textwidth}{!}{
        \setlength{\tabcolsep}{10pt}
        \renewcommand{\arraystretch}{1.3}
        \begin{tabular}{l ccc cc c}
            \toprule
            \multirow{2}{*}{\textbf{Model}} & \multicolumn{3}{c}{\textbf{Attribute Binding} ($\uparrow, \%$)} & \multicolumn{2}{c}{\textbf{Object Relationship} ($\uparrow, \%$)} & \multirow{2}{*}{\textbf{Complex} ($\uparrow, \%$)} \\
            \cmidrule(lr){2-4} \cmidrule(lr){5-6}
            & \textbf{Color} & \textbf{Shape} & \textbf{Texture} & \textbf{Spatial} & \textbf{Non-Spat.} & \\
            \midrule
            \multicolumn{7}{c}{\textit{Unified Multimodal Models}} \\
            \midrule
            Show-o & 56.00 & 41.00 & 46.00 & 20.00 & 30.00 & 29.00 \\
            Janus-Pro-7B & 63.59 & 35.28 & 49.36 & 20.61 & 30.85 & 35.59 \\
            \midrule
            Bagel & 70.62 & 73.55 & 74.21 & 72.15 & 58.20 & 43.10 \\
            \rowcolor{gray!15}
            \quad \textbf{+SRUM} & 72.88 & 76.86 & 77.48 & \textbf{74.40} & \textbf{60.91} & 44.41 \\
            \midrule
            Bagel$_{\text{(CoT)}}$ & 80.95 & 83.92 & 84.35 & 71.80 & 57.55 & 43.65 \\
            \rowcolor{gray!15}
            \quad \textbf{+SRUM} & \textbf{82.14} & \textbf{85.17} & \textbf{89.47} & 74.03 & 59.96 & \textbf{44.51} \\
            \bottomrule
        \end{tabular}
    }
\end{table*}
   
As shown in~\Cref{tab:compbench_final}, our proposed method, \textbf{SRUM}, achieves consistent and substantial performance gains across various compositional generation tasks. Specifically, when evaluated in CoT mode, \textbf{Bagel$_{\text{+SRUM}}$} attains an overall score of 88.37, ranking first among current UMM baselines. This marks a significant improvement of $+$3.91 points over the baseline Bagel with CoT, demonstrating the efficacy of our approach. The advantages of SRUM are particularly pronounced in categories demanding spatial and complex reasoning as well as numeracy. For instance, our method sets new SOTA scores in \textbf{Spatial} (\textbf{93.88}), \textbf{3D Spatial} (\textbf{88.60}) and \textbf{Complex} (\textbf{91.31}) reasoning, including 3D and action-based prompts. Although we observe a slight drop in texture and color categories in some cases, the overall trend remains positive, likely because our algorithm does not over-optimize low-level information for certain objects. Additionally, we report traditional metrics in~\Cref{tab:appendix_traditional_metrics_simplified} to provide a multi-angle comparison. While these evaluators confirm the consistent gains of \textbf{SRUM}, the improvements appear less pronounced than those in the LLM-based evaluation (\Cref{tab:compbench_final}). This discrepancy stems from the inherent limitations of traditional models (e.g., CLIP or BLIP), which often suffer from ``bag-of-words'' effects and lack fine-grained logical parsing. In contrast, our \textbf{Qwen-based evaluation} offers a more solid and nuanced assessment by leveraging superior multimodal reasoning, capturing complex spatial and attribute relationships that traditional metrics tend to overlook.
\vspace{-1em}

\subsection{Empirical Study}
\label{sec:emp}
     
We primarily employ three Bagel variants for analysis: \textbf{Base Model,} where Bagel's open-source weights are used directly for inference; \textbf{SFT Model,} where Bagel generates images from training instructions and is then trained on this self-generated data; and \textbf{SRUM Model,} where the same self-generated data is trained with SRUM's reward-weighted objective.

\textbf{Computational Efficiency.} 
A key advantage of SRUM is its practical efficiency. By parallelizing the understanding module, scoring a batch of 6K candidate images requires fewer than 4 GPU-hours on an NVIDIA H100. Furthermore, the loss computation introduces negligible overhead compared to standard SFT (12.5 GPU-hours for SRUM vs. 12.4 GPU-hours for SFT). Crucially, this self-rewarding mechanism eliminates the need for laborious manual data cleaning, making it highly scalable and practical given the current scarcity of high-quality image-text data.

\begin{figure}[htbp]
    \centering 
    \vspace*{-0.01\textwidth}
    
    \begin{minipage}[t]{0.33\textwidth}
        \centering
        \includegraphics[width=\textwidth]{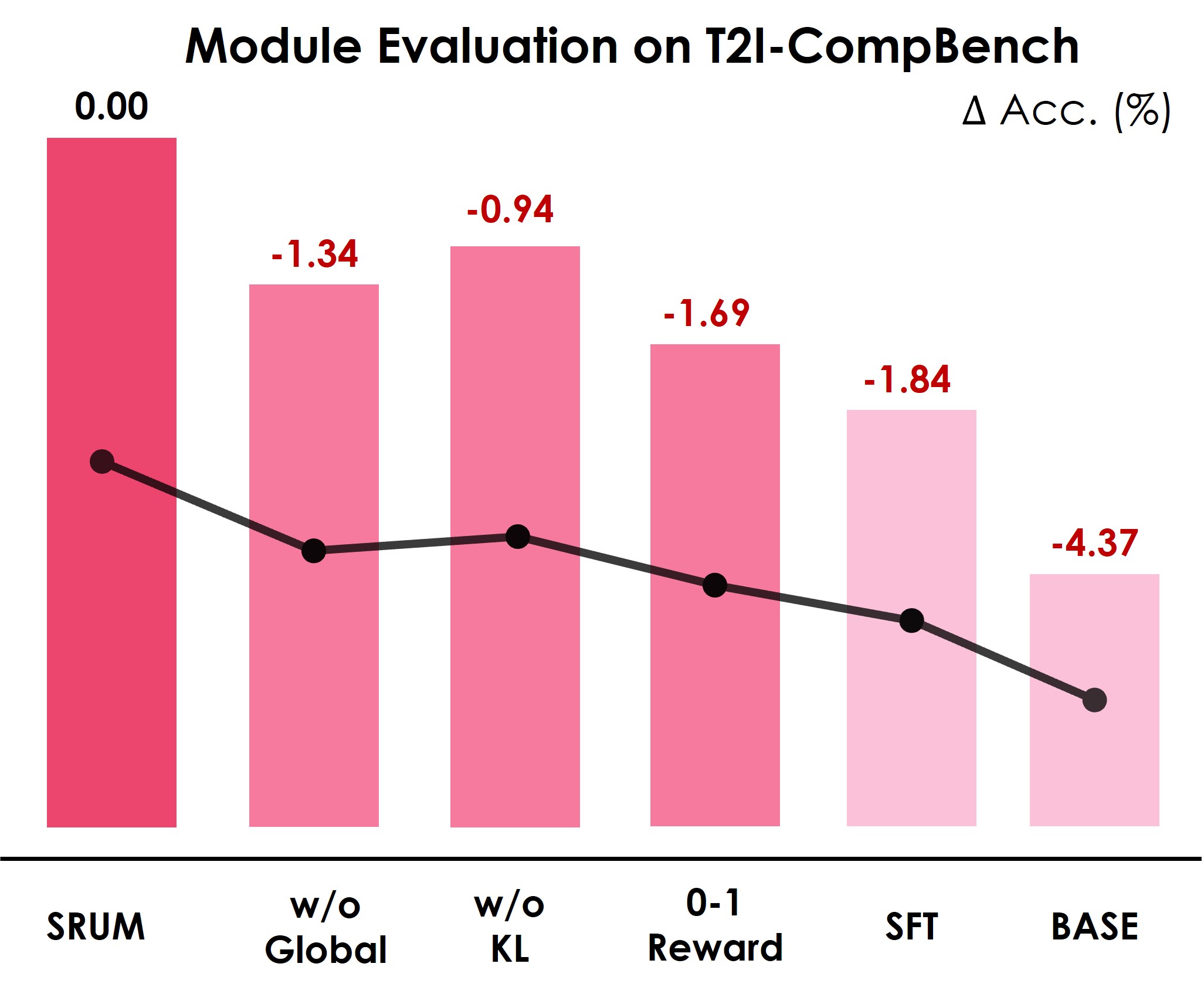}
    \end{minipage}
    \begin{minipage}[t]{0.35\textwidth}
        \centering
        \includegraphics[width=\textwidth]{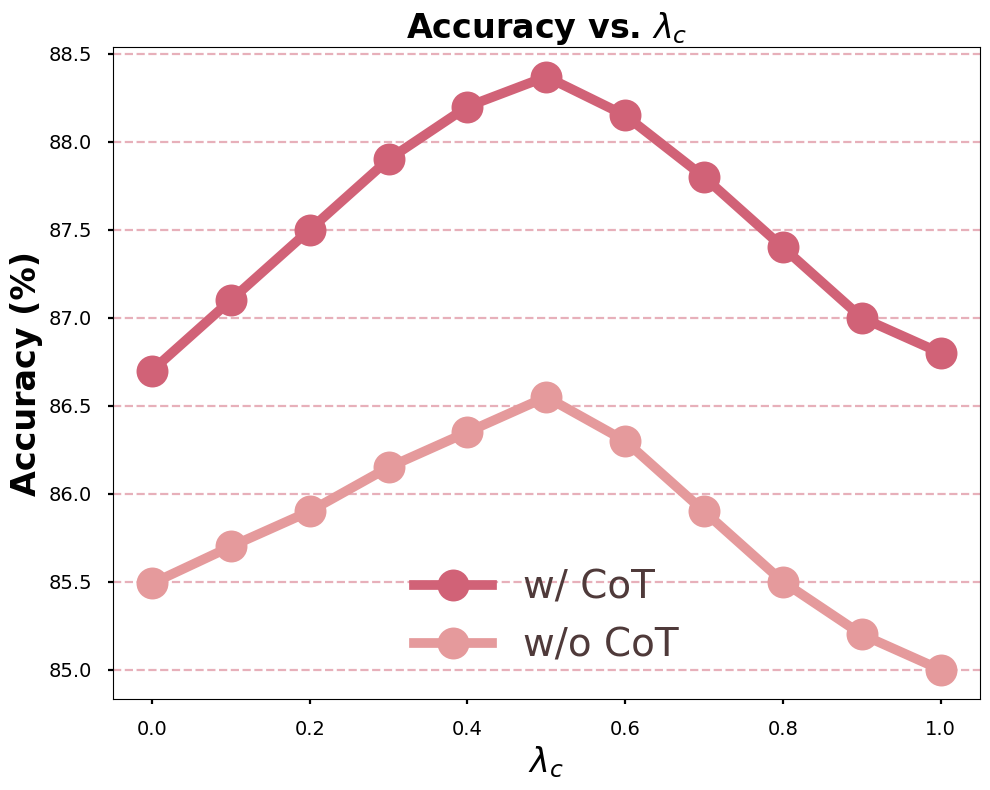}
    \end{minipage}
    
    \vspace{-1em} 
    \caption{
        \textbf{Left:} Module Evaluation. We report the accuracy drop ($\Delta$ Acc. \%) from our SRUM. Specifically, 0-1 Reward represents the sparse reward.
        \textbf{Right:} Hyperparameter evaluation on T2I-CompBench. We report the accuracy under different $\lambda_c$ values in two modes: CoT and non-CoT.
    }
    \label{fig:combined_ab_kl} 
\end{figure}

\textbf{Ablation Results}. To further verify the effectiveness of our proposed reward configuration, we perform an ablation study on Bagel results on T2I-CompBench by systematically modifying the reward scheme. As shown in~\Cref{fig:combined_ab_kl} (Left), our full SRUM model achieves the highest overall accuracy, with the ablation results confirming the critical role of each component. Specifically, omitting the \textbf{local reward} leads to a performance drop of 0.76 in CoT mode and 1.04 in standard inference mode, confirming the necessity of fine-grained spatial feedback that bounding boxes provide. Omitting the \textbf{global reward} leads to a notable decrease in performance, underscoring its importance for capturing the overarching coherence and compositional structure of the generated images. Removing the \textbf{reference constraint} also results in a significant drop, proving its value in ensuring training stability. This aligns with conclusions from post-training methods like DPO~\cite{rafailov2023direct}, where a constraint is essential to prevent significant policy deviation due to reward hacking. Furthermore, using a simple \textbf{sparse reward} leads to substantial performance degradation, reinforcing the necessity of a continuous, dense reward signal for richer gradient information. This is particularly evident as sparse reward schemes, such as Dance-GRPO~\cite{xue2025dancegrpo}, are ill-suited for providing granular regional feedback, which highlights the value of our dense reward design. Overall, this ablation confirms that the efficacy of our framework stems from the synergistic contributions of each component.
\vspace{-1em}

\vspace{-2em}

In~\Cref{fig:combined_ab_kl} (Right), we analyze the effect of different reference-constraint ratios on the experimental outcomes. Across both Bagel with CoT and non-CoT configurations, the results consistently indicate that $\lambda_{c} = 0.5$ is the most effective choice. Consequently, we keep this hyperparameter fixed in subsequent experiments.
 \vspace{-1em}
 
\begin{table}[htbp]
    \centering
    \footnotesize
    \caption{Comparison with external reward models and post-training methods on T2I-CompBench. UR and GR denote UnifiedReward and Global Reward, respectively. Our SRUM achieves the highest overall score.}
    \label{tab:external_reward}
    \resizebox{\columnwidth}{!}{
    \begin{tabular}{lccccccc}
        \toprule
        \textbf{Method} & UR & ReCA & Qwen & Qwen & Qwen & Qwen & \cellcolor{gray!15}\textbf{SRUM (Ours)} \\
        \textbf{Size} & 7B & - & 7B & 7B & 32B & 32B & \cellcolor{gray!15}7B \\
        \midrule
        \textbf{Config.} & - & - & w/o GR & w/ GR & w/o GR & w/ GR & \cellcolor{gray!15}Self-Rewarding \\
        \midrule
        \textbf{Overall Score} & 85.15 & 86.45 & 85.78 & 87.52 & 85.53 & 88.01 & \cellcolor{gray!15}\textbf{88.37} \\
        \bottomrule
    \end{tabular}%
    }
\end{table}
 \vspace{-2em}

\textbf{Comparison with External Rewards and Post-Training Methods.} To further validate the necessity and superiority of the \textit{self-rewarding} mechanism, we compare SRUM against popular post-training baselines including Reconstruction Alignment (ReCA) and the RL-based UnifiedReward, as well as external VLM reward models (QwenVL2.5-7B and 32B). As shown in \Cref{tab:external_reward}, SRUM significantly outperforms ReCA and UnifiedReward. Counter-intuitively, replacing our internal understanding module with powerful external VLMs does not yield stable improvements. On saved image/prompt pairs, the internal scorer achieves stronger calibration with the final evaluator (\(\rho=.58\) vs. \(.43\)) and better accept/reject separation (0.67 vs. -0.31), explaining why it can be a better training signal even when external VLMs are larger. This shows that SRUM's effectiveness comes from the intrinsic multi-scale design rather than external parameter scale.

\textbf{Further Analysis}. For a more granular investigation, we leverage the same powerful MLLM, QwenVL-2.5-72B, from our primary evaluation to conduct a deeper analysis of our method and the baseline. Specifically, we employ the MLLM to perform a step-by-step scoring of the inference process.
\vspace{-1em}

\begin{figure*}[!htbp]
\vspace{-0.02\textwidth}
\centering
\begin{subfigure}[b]{0.88\textwidth}
    \centering
    \includegraphics[width=\textwidth]{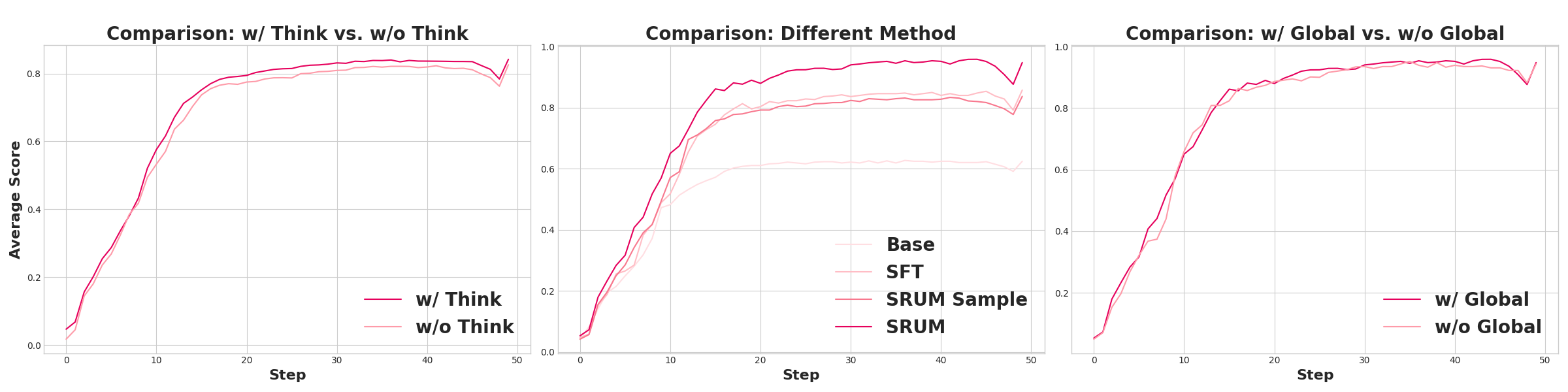}
    \vspace{-1.5em}
    \caption{\scriptsize{\textbf{Detail score per step during inference}}}
    \label{fig:plot_detail_score_comparisons_absolute}
\end{subfigure}

\vspace{0.5em}

\begin{subfigure}[b]{0.88\textwidth}
    \centering
    \includegraphics[width=\textwidth]{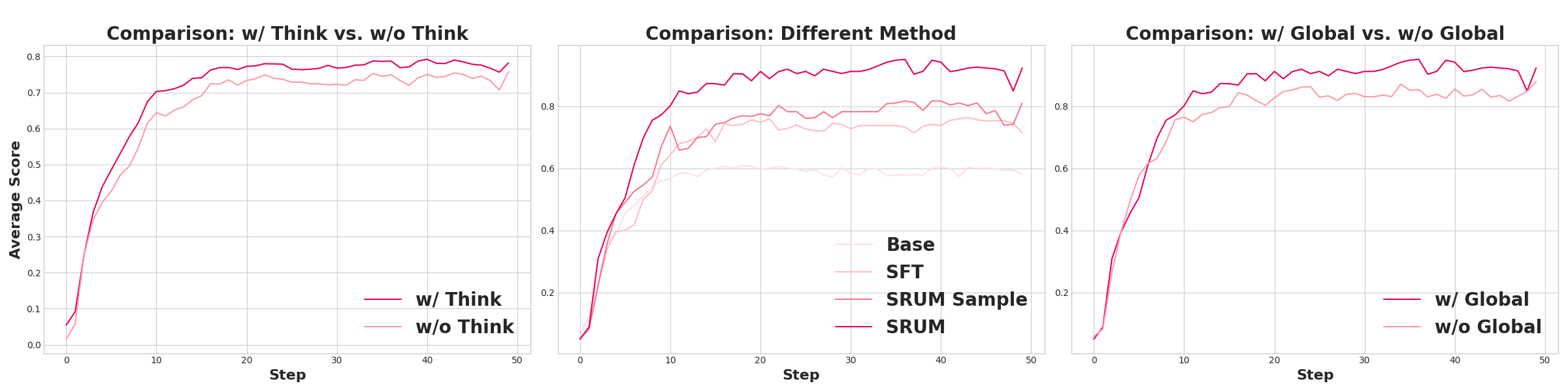}
    \caption{\scriptsize{\textbf{Layout score per step during inference}}}
    \label{fig:plot_layout_score_comparisons_absolute}
\end{subfigure}
\vspace{0.5em}
      \caption{Score per step during inference in Bagel with its ablation models.}
\vspace{-1em}
\label{fig:score}
\end{figure*}
\vspace{-1em}

The evaluation is divided into two metrics: (1) layout, which assesses the concordance of the overall structure and quality, and (2) detail, which measures the fidelity of the generated fine-grained details. Our ablation study, visualized in~\Cref{fig:score}, systematically isolates the effects of each component. We observe that the ``think'' mode primarily bolsters the initial layout generation by improving the high-level reasoning process. The global reward component of SRUM then further refines this layout during the early stages of inference. In contrast, a baseline using only this global reward (labeled ``sample reward'') yields negligible improvements in detail fidelity. This highlights a crucial finding: the fine-grained, local rewards are essential for the subsequent optimization of details, with their benefits becoming most apparent in the later inference steps. Collectively, these results demonstrate that our dual global-local reward mechanism provides a multi-stage optimization path: first establishing a coherent layout and then progressively refining the details. This synergistic approach allows SRUM to significantly outperform standard SFT on the same self-generated data.

\textbf{Impact on Understanding Module}. As shown in \Cref{tab:vlm_eval}, our method has a minimal impact on the model's core understanding capabilities. On prevalent benchmarks such as MME~\cite{fu2023mme}, MM-Vet~\cite{yu2024mm}, MMBench~\cite{liu2024mmbench}, MMMU~\cite{yue2024mmmu}, and MathVista~\cite{lu23mathvista}, the results exhibit only marginal fluctuations compared to the base version. Notably, performance on MMVP~\cite{tong2024eyes} even improves, which is consistent with prior works~\cite{tong2024metamorph,wang2024diffusion,wang2024reconstructive}. This indicates that our method holds significant potential for further iterative enhancement. In~\Cref{fig:activation}, we track the activation dynamics of two distinct functional clusters, Understanding and Generation, across the Base, SFT, and SRUM models. In Bagel's inference, the parameters primarily activated during understanding tasks are defined as the understanding cluster, and those primarily activated during generation tasks are defined as the generation cluster. Our analysis reveals two distinct finetuning paradigms. Conventional SFT exhibits a narrowing effect, achieving specialization by suppressing irrelevant functional clusters. In contrast, SRUM demonstrates an enhancing and orchestrating effect, strengthening the primary task-relevant cluster while maintaining supportive activation in secondary clusters. This promotes robust and generalizable representations. Detailed settings are provided in Appendix~\ref{sec:appendix_activation}.
\begin{center}
    \begin{minipage}[c]{0.504\textwidth}
            \vspace{-0.5em}
        \centering
        \includegraphics[width=0.99\textwidth]{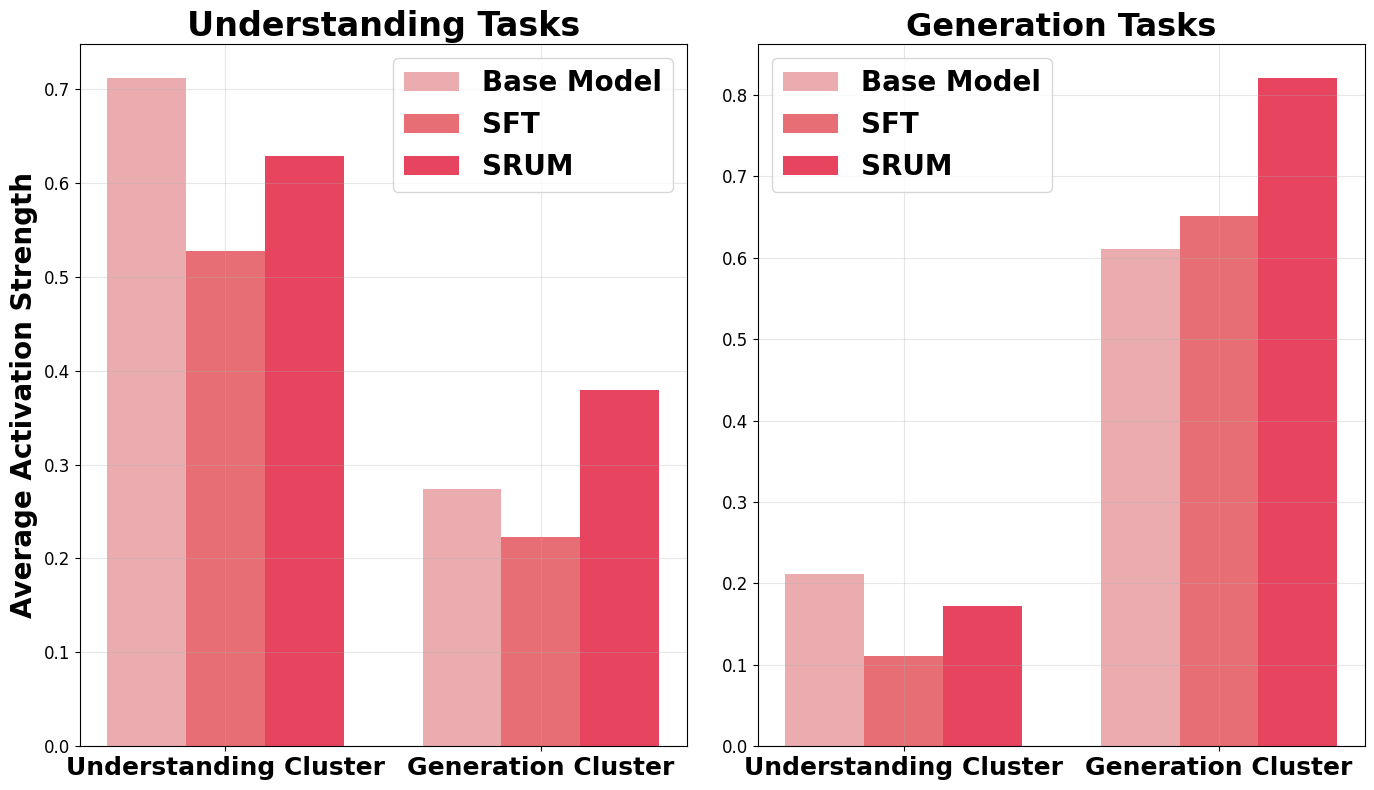}
        \captionof{figure}{Functional cluster activation patterns of the different models (Bagel, SFT and SRUM) on different understanding and generation tasks. The average activation strength of Understanding and Generation clusters is shown.}
        \label{fig:activation}
    \end{minipage}
    \hspace{0.7em}
    \begin{minipage}[c]{0.41\textwidth}
                \vspace{-0.5em}
        \centering
        \small
        \begin{tabular}{lccc}
            \toprule
            & \textbf{Base} & \textbf{SFT} & \textbf{SRUM} \\
            \midrule
            \textbf{MME-P} & 1687 & 1682 & 1673 \\
            \textbf{MME-C} & 701 & 683 & 677 \\
            \textbf{MMBench} & 85.0 & 84.6 & 84.8 \\
            \textbf{MM-Vet} & 67.2 & 66.5 & 67.0 \\
            \textbf{MMMU} & 55.3 & 55.0 & 55.2 \\
            \textbf{MathVista} & 73.1 & 72.8 & 73.0 \\
            \textbf{MMVP} & 69.3 & 68.7 & 70.0 \\
            \bottomrule
        \end{tabular}
        \vspace{0.5em}
        \captionof{table}{Comparison with the results of different models (Bagel, SFT and SRUM) on understanding benchmarks. MME-P and MME-C represent the perception and cognition parts, respectively.}
        \label{tab:vlm_eval}
    \end{minipage}
\end{center}

\textbf{In-Domain Generalization}. We then investigate the in-domain generalization capability of our model. We posit that the compositional abilities learned from the T2I-CompBench training set should be transferable to other benchmarks with similar evaluation perspectives. To test this hypothesis, we evaluate SRUM on the GenEval benchmark without any further fine-tuning. The comparative results are summarized in \Cref{tab:ablation_attributes}. 

As shown in the table, our evaluation on GenEval further validates the strengths of SRUM, particularly in the challenging domain of object counting. SRUM attains the highest score of 0.83 in Counting, surpassing both the base model and the SFT baseline. Crucially, this superior performance in numerical generation aligns with our previous results on T2I-CompBench. This consistency across benchmarks underscores our method's reliable improvement in processing quantitative information. By excelling at a complex task like counting while retaining proficiency in simpler ones, the model demonstrates strong in-domain generalization. This confirms that the targeted enhancements by SRUM are transferable.
\vspace{-2em}

\begin{table}[htbp]
    \centering
    \small
    \caption{Results on key visual attributes at GenEval. For brevity, some model names are shortened; abbreviations are defined at the beginning of~\Cref{sec:emp}. \textbf{Bold values} are the best in each column.}
    \label{tab:ablation_attributes}
    \begin{tabular}{lcccccc}
        \toprule
        \textbf{Model} & {\textbf{Single obj.}} & {\textbf{Two obj.}} & {\textbf{Counting}} & {\textbf{Colors}} & {\textbf{Position}} & {\textbf{Color attr.}} \\
        \midrule
        Bagel & \bfseries 0.99 & 0.94 & 0.81 & 0.88 & 0.64 & 0.82 \\
        Bagel$_{\text{+SFT}}$ & 0.96 & 0.94 & 0.79 &\bfseries 0.92 & 0.59 & 0.78 \\
        \rowcolor{gray!15}
        Bagel$_{\text{+SRUM}}$ & 0.98 & 0.94 &\bfseries 0.83 & 0.90 & 0.64 & \bfseries 0.83 \\
        \bottomrule
    \end{tabular}
      \vspace{-0.5em}
\end{table}
\vspace{-1em}

\textbf{In-Domain Knowledge-based Generalization}. Following this, we explore whether our method holds a distinct advantage for reasoning-based generation, a current area of focus in the community. We train the model on one category of prompts from the WISE benchmark and perform in-domain evaluations on the remaining two categories. This protocol yields three distinct evaluation sets for analyzing the model's generalization capabilities.

\begin{figure}[htbp]  
  \hspace*{-0.02\textwidth}
  \includegraphics[width=0.75\textwidth]{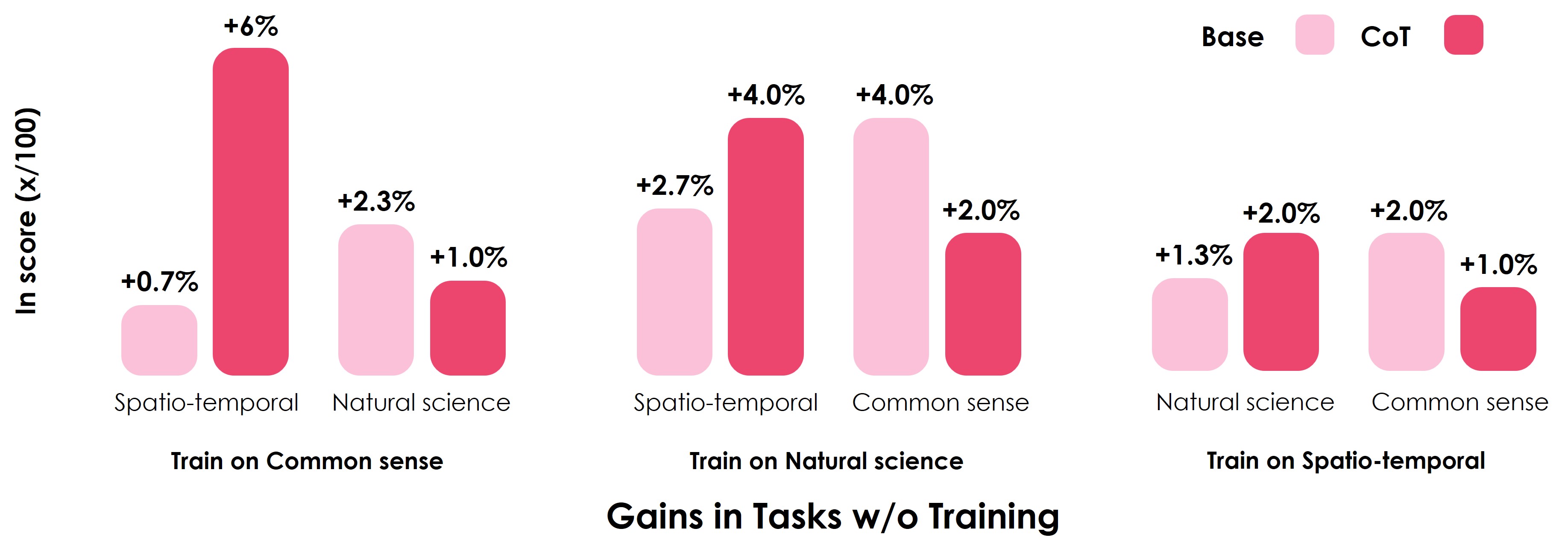}
       \vspace{-1em}
  \caption{Results of Bagel on WISE. We train on one of the three WISE tasks and evaluate on the other two, showing SRUM's knowledge-domain generalization.}
  \label{fig:wise}
\end{figure}

As illustrated in~\Cref{fig:wise}, selecting any single group for training generally enhances the image generation performance of the other two groups. This improvement is consistent across both standard and CoT reasoning paradigms. It shows that SRUM can promote knowledge-domain generalization, enabling generation to better fit instruction semantics at the knowledge level.
\vspace{-1em}

\textbf{Out-of-Domain Generalization}. To further evaluate the \textbf{generalization capability} of our model on unseen domains, we use \textbf{T2I-ReasonBench}, a large-scale benchmark for analyzing the reasoning quality of generated images. In this experiment, we take the model trained with \textbf{T2I-CompBench} prompts and directly evaluate its performance on this benchmark. This setup demonstrates the model's ability to generalize to advanced, reasoning-informed image generation tasks. We primarily focus on accuracy scores, which measure the model's high-level semantic alignment with the given prompts. To prevent self-rewrite from directly parsing hidden high-level semantics (e.g., in the \textbf{Idiom} category, a phrase like ``a piece of cake'' might be literally interpreted as ``easy,'' which would obscure the model's ability to transfer understanding and could interfere with evaluation), we use Bagel without CoT during evaluation.

As illustrated in~\Cref{tab:reason}, SRUM achieves superior prompt understanding compared with both the SFT and Base models. While SFT also yields a noticeable improvement, the enhanced performance of SRUM demonstrates that our approach improves generalization on complex problems from both a data and an algorithmic perspective. Furthermore, for image-based prompts, SRUM provides consistent improvements, in contrast to the volatility exhibited by the SFT model. This further substantiates that our algorithmic design is more adaptable and accounts for more nuanced factors. Finally, visual case studies and a detailed analysis of failure modes are provided in Appendix~\ref{fail}, helping differentiate algorithmic boundaries from inherent generative model limitations.
\vspace{-1em}

\begin{table}[htbp]
    \centering
    \begin{tabular}{l ccccc}
        \toprule
        \textbf{Model} & \textbf{Entity} & \textbf{Idiom} & \textbf{Scientific} & \textbf{Textual} & \textbf{Overall} \\
        \midrule
        Bagel & 49.70 & 34.46 & 47.52 & 43.59 & 43.82 \\
        Bagel$_{\text{+SFT}}$ & 50.53 & 39.43 & 47.45 & 44.08 & 45.37 \\
        \rowcolor{gray!15}
        Bagel$_{\text{+SRUM}}$ & \bfseries 52.85 & \bfseries 40.51 & \bfseries 47.83 & \bfseries 45.83 & \bfseries 46.75 \\
        \bottomrule
    \end{tabular}
    \caption{Performance comparison of Bagel models across four categories and their overall scores. \textbf{Bold values} indicate the best performance in each column. Scores are normalized between 0-100.}
    \label{tab:reason}
\end{table}
\vspace{-3em}
\section{Conclusion}
\label{sec:conclusion} 
This paper introduces SRUM, a fine-grained post-training framework that enables a model's understanding module to reward its generation module. Additionally, SRUM decomposes the reward into local and global components, facilitating multi-scale alignment and refinement. Extensive experiments validate SRUM's effectiveness, setting new state-of-the-art results on complex compositional and reasoning benchmarks such as T2I-CompBench and T2I-ReasonBench. The framework demonstrates robust in-domain and out-of-domain generalization, and our empirical analysis confirms the efficacy of the fine-grained reward design. These findings illuminate the synergistic development of understanding and generation capabilities within a single model and establish the principle of self-reward as a promising direction for future research.

SRUM is a preliminary exploration of self-rewarding for Unified Multimodal Models (UMMs). There remains room to improve the prompts used by the understanding module during scoring, and we plan to scale the method to larger datasets. Although the current implementation uses standardized prompt templates and, for some models, external localization aids, the reward signal itself comes from the UMM's frozen understanding module. A natural next step is to let the understanding module \textbf{self-play questions and answers} to build a more closed-loop training system.

\section*{Acknowledgements}
This work is supported by the Shenzhen Loop Area Institute under grant FPF \\ 10120250006, and partially supporteed under grants FPFFPF10120250002.

\bibliographystyle{splncs04}
\bibliography{main}

@String(CVPR= {IEEE Conf. Comput. Vis. Pattern Recog.})

@String(ECCV= {Eur. Conf. Comput. Vis.})

@String(ICLR = {Int. Conf. Learn. Represent.})

@String(CVPR  = {CVPR})

@String(ECCV  = {ECCV})

@String(ICLR  = {ICLR})

@article{gemini,
  title={Gemini: a family of highly capable multimodal models},
  author={Team, Gemini and Anil, Rohan and Borgeaud, Sebastian and Alayrac, Jean-Baptiste and Yu, Jiahui and Soricut, Radu and Schalkwyk, Johan and Dai, Andrew M and Hauth, Anja and Millican, Katie and others},
  journal={arXiv preprint arxiv:2312.11805},
  year={2023}
}

@inproceedings{sdxl,
  title={Sdxl: Improving latent diffusion models for high-resolution image synthesis},
  author={Podell, Dustin and English, Zion and Lacey, Kyle and Blattmann, Andreas and Dockhorn, Tim and M{\"u}ller, Jonas and Penna, Joe and Rombach, Robin},
  booktitle = {ICLR},
  year = {2024}
}

@inproceedings{dalle1,
  title={Zero-Shot Text-to-Image Generation},
  author={Ramesh, Aditya and Pavlov, Mikhail and Goh, Gabriel and Gray, Scott and Voss, Chelsea and Radford, Alec and Chen, Mark and Sutskever, Ilya},
  booktitle={ICML},
  year={2021}
}

@article{dalle3,
  title={Improving image generation with better captions},
  author={Betker, James and Goh, Gabriel and Jing, Li and Brooks, Tim and Wang, Jianfeng and Li, Linjie and Ouyang, Long and Zhuang, Juntang and Lee, Joyce and Guo, Yufei and others},
  journal={OpenAI blog},
  year={2023}
}

@inproceedings{song1,
  title={Score-based generative modeling through stochastic differential equations},
  author={Song, Yang and Sohl-Dickstein, Jascha and Kingma, Diederik P and Kumar, Abhishek and Ermon, Stefano and Poole, Ben},
  booktitle={ICLR},
  year={2021}
}

@inproceedings{imagen,
  title={Photorealistic text-to-image diffusion models with deep language understanding},
  author={Saharia, Chitwan and Chan, William and Saxena, Saurabh and Li, Lala and Whang, Jay and Denton, Emily L and Ghasemipour, Kamyar and Gontijo Lopes, Raphael and Karagol Ayan, Burcu and Salimans, Tim and others},
  booktitle={NeurIPS},
  year={2022}
}

@misc{flux,
  title = {FLUX},
  author = {Black Forest Labs},
  year = {2024},
  url = {https://github.com/black-forest-labs/flux}
}

@inproceedings{transformer,
  title={Attention Is All You Need}, 
  author={Ashish Vaswani and Noam Shazeer and Niki Parmar and Jakob Uszkoreit and Llion Jones and Aidan N. Gomez and Lukasz Kaiser and Illia Polosukhin},
  booktitle={NeurIPS},
  year={2017}
}

@article{emu3,
  title={Emu3: Next-token prediction is all you need},
  author={Wang, Xinlong and Zhang, Xiaosong and Luo, Zhengxiong and Sun, Quan and Cui, Yufeng and Wang, Jinsheng and Zhang, Fan and Wang, Yueze and Li, Zhen and Yu, Qiying and others},
  journal={arXiv preprint arxiv:2409.18869},
  year={2024}
}

@inproceedings{llava-plus,
  title={Llava-plus: Learning to use tools for creating multimodal agents},
  author={Liu, Shilong and Cheng, Hao and Liu, Haotian and Zhang, Hao and Li, Feng and Ren, Tianhe and Zou, Xueyan and Yang, Jianwei and Su, Hang and Zhu, Jun and others},
  booktitle={ECCV},
  year={2024}
}

@article{chameleon,
  title={Chameleon: Mixed-modal early-fusion foundation models},
  author={Team, Chameleon},
  journal={arXiv preprint arXiv:2405.09818},
  year={2024}
}

@article{show-o,
  title={Show-o: One single transformer to unify multimodal understanding and generation},
  author={Xie, Jinheng and Mao, Weijia and Bai, Zechen and Zhang, David Junhao and Wang, Weihao and Lin, Kevin Qinghong and Gu, Yuchao and Chen, Zhijie and Yang, Zhenheng and Shou, Mike Zheng},
  journal={arXiv preprint arxiv:2408.12528},
  year={2024}
}

@inproceedings{ddpm,
  title={Denoising diffusion probabilistic models},
  author={Ho, Jonathan and Jain, Ajay and Abbeel, Pieter},
  booktitle={NeurIPS},
  year={2020}
}

@article{qwen2vl,
  title={Qwen2-VL: Enhancing Vision-Language Model's Perception of the World at Any Resolution}, 
  author={Peng Wang and Shuai Bai and Sinan Tan and Shijie Wang and Zhihao Fan and Jinze Bai and Keqin Chen and Xuejing Liu and Jialin Wang and Wenbin Ge and Yang Fan and Kai Dang and Mengfei Du and Xuancheng Ren and Rui Men and Dayiheng Liu and Chang Zhou and Jingren Zhou and Junyang Lin},
  journal={arXiv preprint arXiv:2409.12191},
  year={2024},
}

@inproceedings{sohl2015deep,
  title={Deep unsupervised learning using nonequilibrium thermodynamics},
  author={Sohl-Dickstein, Jascha and Weiss, Eric and Maheswaranathan, Niru and Ganguli, Surya},
  booktitle={ICML},
  pages={2256--2265},
  year={2015},
  organization={PMLR}
}

@article{hoffmann2022training,
  title={Training Compute-Optimal Large Language Models},
  author={Hoffmann, Jordan and Borgeaud, Sebastian and Mensch, Arthur and Buchatskaya, Elena and Cai, Trevor and Rutherford, Eliza and de Las Casas, Diego and Hendricks, Lisa Anne and Welbl, Johannes and Clark, Aidan and Hennigan, Tom and Noland, Eric and Millican, Katie and van den Driessche, George and Damoc, Bogdan and Guy, Aurelia and Osindero, Simon and Simonyan, Karen and Elsen, Erich and Rae, Jack W. and Vinyals, Oriol and Sifre, Laurent},
  journal={arXiv preprint arxiv:2203.15556},
  year={2022}
}

@inproceedings{kaplan2020scaling,
  title={Scaling Laws for Neural Language Models},
  author={Kaplan, Jared and McCandlish, Sam and Henighan, Tom and Brown, Tom and Chess, Benjamin and Child, Rewon and Gray, Scott and Radford, Alec and Wu, Jeffrey and Amodei, Dario},
  booktitle={ICML},
  year={2020},
}

@inproceedings{vit,
  title={An Image is Worth 16x16 Words: Transformers for Image Recognition at Scale},
  author={Alexey Dosovitskiy and Lucas Beyer and Alexander Kolesnikov and Dirk Weissenborn and Xiaohua Zhai and Thomas Unterthiner and Mostafa Dehghani and Matthias Minderer and Georg Heigold and Sylvain Gelly and Jakob Uszkoreit and Neil Houlsby},
  booktitle={ICLR},
  year={2021},
}

@inproceedings{SD3,
  title={Scaling rectified flow transformers for high-resolution image synthesis},
  author={Esser, Patrick and Kulal, Sumith and Blattmann, Andreas and Entezari, Rahim and M{\"u}ller, Jonas and Saini, Harry and Levi, Yam and Lorenz, Dominik and Sauer, Axel and Boesel, Frederic and others},
  booktitle={ICML},
  year={2024}
}

@inproceedings{rombach2022high,
  title={High-resolution image synthesis with latent diffusion models},
  author={Rombach, Robin and Blattmann, Andreas and Lorenz, Dominik and Esser, Patrick and Ommer, Bj{\"o}rn},
  booktitle={CVPR},
  pages={10684--10695},
  year={2022}
}

@article{goyal2017accurate,
  title={Accurate, large minibatch SG D: training imagenet in 1 hour},
  author={Goyal, P},
  journal={arXiv preprint arXiv:1706.02677},
  year={2017}
}

@article{ho2022classifier,
  title={Classifier-free diffusion guidance},
  author={Ho, Jonathan and Salimans, Tim},
  journal={arXiv preprint arXiv:2207.12598},
  year={2022}
}

@inproceedings{lin2014microsoft,
  title={Microsoft coco: Common objects in context},
  author={Lin, Tsung-Yi and Maire, Michael and Belongie, Serge and Hays, James and Perona, Pietro and Ramanan, Deva and Doll{\'a}r, Piotr and Zitnick, C Lawrence},
  booktitle={ECCV},
  year={2014},
}

@article{loshchilov2017decoupled,
  title={Decoupled weight decay regularization},
  author={Loshchilov, I},
  journal={arXiv preprint arXiv:1711.05101},
  year={2017}
}

@inproceedings{chen2024diffusion,
  title={Diffusion forcing: Next-token prediction meets full-sequence diffusion},
  author={Chen, Boyuan and Mart{\'\i} Mons{\'o}, Diego and Du, Yilun and Simchowitz, Max and Tedrake, Russ and Sitzmann, Vincent},
  booktitle={NeurIPS},
  year={2024}
}

@article{wang2025mint,
  title={MINT: Multi-modal Chain of Thought in Unified Generative Models for Enhanced Image Generation},
  author={Wang, Yi and Liu, Mushui and He, Wanggui and Zhang, Longxiang and Huang, Ziwei and Zhang, Guanghao and Shu, Fangxun and Tao, Zhong and She, Dong and Yu, Zhelun and others},
  journal={arXiv preprint arXiv:2503.01298},
  year={2025}
}

@article{shi2024llamafusion,
  title={LlamaFusion: Adapting Pretrained Language Models for Multimodal Generation},
  author={Shi, Weijia and Han, Xiaochuang and Zhou, Chunting and Liang, Weixin and Lin, Xi Victoria and Zettlemoyer, Luke and Yu, Lili},
  journal={arXiv preprint arXiv:2412.15188},
  year={2024}
}

@article{januspro2025,
  title={Janus-Pro: Unified Multimodal Understanding and Generation with Data and Model Scaling},
  author={Xiaokang Chen and Chengyue Wu and Zhiyu Wu and Yiyang Ma and Xingchao Liu and Zizheng Pan and Wen Liu and Zhenda Xie and Xingkai Yu and Chong Ruan and Ping Luo},
  journal={arXiv preprint arXiv:2501.17811},
  year={2025},
}

@article{monoformer2024,
  title={MonoFormer: One Transformer for Both Diffusion and Autoregression},
  author={Chuyang Zhao and Yuxing Song and Wenhao Wang and Haocheng Feng and Errui Ding and Yifan Sun and Xinyan Xiao and Jingdong Wang},
  journal={arXiv preprint arXiv:2409.16280},
  year={2024},
  url={https://arxiv.org/abs/2409.16280}
}

@article{janus2024,
  title={Janus: Decoupling Visual Encoding for Unified Multimodal Understanding and Generation},
  author={Chengyue Wu and Xiaokang Chen and Zhiyu Wu and Yiyang Ma and Xingchao Liu and Zizheng Pan and Wen Liu and Zhenda Xie and Xingkai Yu and Chong Ruan and Ping Luo},
  journal={arXiv preprint arXiv:2410.13848},
  year={2024},
}

@article{vilau2024,
  title={VILA-U: A Unified Foundation Model Integrating Visual Understanding and Generation},
  author={Yecheng Wu and Zhuoyang Zhang and Junyu Chen and Haotian Tang and Dacheng Li and Yunhao Fang and Ligeng Zhu and Enze Xie and Hongxu Yin and Li Yi and Song Han and Yao Lu},
  journal={arXiv preprint arXiv:2409.04429},
  year={2024},
  url={https://arxiv.org/abs/2409.04429}
}

@article{tong2024metamorph,
  title={MetaMorph: Multimodal Understanding and Generation via Instruction Tuning},
  author={Shengbang Tong and David Fan and Jiachen Zhu and Yunyang Xiong and Xinlei Chen and Koustuv Sinha and Michael Rabbat and Yann LeCun and Saining Xie and Zhuang Liu},
  journal={arXiv preprint arXiv:2412.14164},
  year={2024},
}

@article{qwen2.5-vl,
      title={Qwen2.5-VL Technical Report}, 
      author={Shuai Bai and Keqin Chen and Xuejing Liu and Jialin Wang and Wenbin Ge and Sibo Song and Kai Dang and Peng Wang and Shijie Wang and Jun Tang and Humen Zhong and Yuanzhi Zhu and Mingkun Yang and Zhaohai Li and Jianqiang Wan and Pengfei Wang and Wei Ding and Zheren Fu and Yiheng Xu and Jiabo Ye and Xi Zhang and Tianbao Xie and Zesen Cheng and Hang Zhang and Zhibo Yang and Haiyang Xu and Junyang Lin},
      year={2025},
      journal={arXiv preprint arXiv:2502.13923},
}

@article{internvl2,
  title={How far are we to gpt-4v? closing the gap to commercial multimodal models with open-source suites},
  author={Chen, Zhe and Wang, Weiyun and Tian, Hao and Ye, Shenglong and Gao, Zhangwei and Cui, Erfei and Tong, Wenwen and Hu, Kongzhi and Luo, Jiapeng and Ma, Zheng and others},
  journal={SCIS},
  year={2024},
}

@inproceedings{wei2024omniedit,
  title={Omniedit: Building image editing generalist models through specialist supervision},
  author={Wei, Cong and Xiong, Zheyang and Ren, Weiming and Du, Xeron and Zhang, Ge and Chen, Wenhu},
  booktitle={ICLR},
  year={2024}
}

@inproceedings{zhang2023magicbrush,
  title={Magicbrush: A manually annotated dataset for instruction-guided image editing},
  author={Zhang, Kai and Mo, Lingbo and Chen, Wenhu and Sun, Huan and Su, Yu},
  booktitle={NeurIPS},
  year={2023}
}

@article{hui2024hqedit,
  title={Hq-edit: A high-quality dataset for instruction-based image editing},
  author={Hui, Mude and Yang, Siwei and Zhao, Bingchen and Shi, Yichun and Wang, Heng and Wang, Peng and Zhou, Yuyin and Xie, Cihang},
  journal={arXiv preprint arXiv:2404.09990},
  year={2024}
}

@article{bai2024humanedit,
  title={HumanEdit: A High-Quality Human-Rewarded Dataset for Instruction-based Image Editing},
  author={Bai, Jinbin and Chow, Wei and Yang, Ling and Li, Xiangtai and Li, Juncheng and Zhang, Hanwang and Yan, Shuicheng},
  journal={arXiv preprint arXiv:2412.04280},
  year={2024}
}

@article{li2024omnicorpus,
  title={OmniCorpus: A Unified Multimodal Corpus of 10 Billion-Level Images Interleaved with Text},
  author={Li, Qingyun and Chen, Zhe and Wang, Weiyun and Wang, Wenhai and Ye, Shenglong and Jin, Zhenjiang and Chen, Guanzhou and He, Yinan and Gao, Zhangwei and Cui, Erfei and others},
  journal={arXiv preprint arXiv:2406.08418},
  year={2024}
}

@article{guo2025r1,
  title={Deepseek-r1: Incentivizing reasoning capability in llms via reinforcement learning},
  author={Guo, Daya and Yang, Dejian and Zhang, Haowei and Song, Junxiao and Zhang, Ruoyu and Xu, Runxin and Zhu, Qihao and Ma, Shirong and Wang, Peiyi and Bi, Xiao and others},
  journal={arXiv preprint arXiv:2501.12948},
  year={2025}
}

@inproceedings{schuhmann2022laion,
  title={Laion-5b: An open large-scale dataset for training next generation image-text models},
  author={Schuhmann, Christoph and Beaumont, Romain and Vencu, Richard and Gordon, Cade and Wightman, Ross and Cherti, Mehdi and Coombes, Theo and Katta, Aarush and Mullis, Clayton and Wortsman, Mitchell and others},
  booktitle={NeurIPS},
  year={2022}
}

@article{li2024datacomprecap,
  title={What If We Recaption Billions of Web Images with LLaMA-3?},
  author={Li, Xianhang and Tu, Haoqin and Hui, Mude and Wang, Zeyu and Zhao, Bingchen and Xiao, Junfei and Ren, Sucheng and Mei, Jieru and Liu, Qing and Zheng, Huangjie and others},
  journal={arXiv preprint arXiv:2406.08478},
  year={2024}
}

@inproceedings{sharma2018cc,
  title={Conceptual captions: A cleaned, hypernymed, image alt-text dataset for automatic image captioning},
  author={Sharma, Piyush and Ding, Nan and Goodman, Sebastian and Soricut, Radu},
  booktitle={ACL},
  year={2018}
}

@inproceedings{lipman2022flow,
  title={Flow matching for generative modeling},
  author={Lipman, Yaron and Chen, Ricky TQ and Ben-Hamu, Heli and Nickel, Maximilian and Le, Matt},
  booktitle={ICLR},
  year={2023}
}

@article{pan2025transfer,
  title={Transfer between Modalities with MetaQueries},
  author={Pan, Xichen and Shukla, Satya Narayan and Singh, Aashu and Zhao, Zhuokai and Mishra, Shlok Kumar and Wang, Jialiang and Xu, Zhiyang and Chen, Jiuhai and Li, Kunpeng and Juefei-Xu, Felix and Hou, Ji and Xie, Saining},
  year={2025},
  journal={arXiv preprint arXiv:2504.06256}
}

@inproceedings{zhang2019root,
  title={Root mean square layer normalization},
  author={Zhang, Biao and Sennrich, Rico},
  booktitle={NeurIPS},
  year={2019}
}

@article{shazeer2020glu,
  title={Glu variants improve transformer},
  author={Shazeer, Noam},
  journal={arXiv preprint arXiv:2002.05202},
  year={2020}
}

@article{su2024roformer,
  title={Roformer: Enhanced transformer with rotary position embedding},
  author={Su, Jianlin and Ahmed, Murtadha and Lu, Yu and Pan, Shengfeng and Bo, Wen and Liu, Yunfeng},
  journal={Neurocomputing},
  volume={568},
  pages={127063},
  year={2024},
  publisher={Elsevier}
}

@article{ainslie2023gqa,
  title={Gqa: Training generalized multi-query transformer models from multi-head checkpoints},
  author={Ainslie, Joshua and Lee-Thorp, James and De Jong, Michiel and Zemlyanskiy, Yury and Lebr{\'o}n, Federico and Sanghai, Sumit},
  journal={arXiv preprint arXiv:2305.13245},
  year={2023}
}

@inproceedings{dongdreamllm,
  title={DreamLLM: Synergistic Multimodal Comprehension and Creation},
  author={Dong, Runpei and Han, Chunrui and Peng, Yuang and Qi, Zekun and Ge, Zheng and Yang, Jinrong and Zhao, Liang and Sun, Jianjian and Zhou, Hongyu and Wei, Haoran and others},
  year={2024},
  booktitle={ICLR}
}

@article{liu2023visual,
  title={Visual instruction tuning},
  author={Liu, Haotian and Li, Chunyuan and Wu, Qingyang and Lee, Yong Jae},
  journal={NeurIPS},
  volume={36},
  pages={34892--34916},
  year={2023}
}

@article{niu2025wise,
  title={Wise: A world knowledge-informed semantic evaluation for text-to-image generation},
  author={Niu, Yuwei and Ning, Munan and Zheng, Mengren and Lin, Bin and Jin, Peng and Liao, Jiaqi and Ning, Kunpeng and Zhu, Bin and Yuan, Li},
  journal={arXiv preprint arXiv:2503.07265},
  year={2025}
}

@inproceedings{liuflow,
  title={Flow Straight and Fast: Learning to Generate and Transfer Data with Rectified Flow},
  author={Liu, Xingchao and Gong, Chengyue and others},
  booktitle={ICLR},
  year={2023},
}

@inproceedings{
    hu2024minicpm,
    title={MiniCPM: Unveiling the Potential of Small Language Models with Scalable Training Strategies},
    author={Shengding Hu and Yuge Tu and Xu Han and Ganqu Cui and Chaoqun He and Weilin Zhao and Xiang Long and Zhi Zheng and Yewei Fang and Yuxiang Huang and Xinrong Zhang and Zhen Leng Thai and Chongyi Wang and Yuan Yao and Chenyang Zhao and Jie Zhou and Jie Cai and Zhongwu Zhai and Ning Ding and Chao Jia and Guoyang Zeng and dahai li and Zhiyuan Liu and Maosong Sun},
    booktitle={COLM},
    year={2024},
}

@article{yu2024anyedit,
    title={AnyEdit: Mastering Unified High-Quality Image Editing for Any Idea},
    author={Yu, Qifan and Chow, Wei and Yue, Zhongqi and Pan, Kaihang and Wu, Yang and Wan, Xiaoyang and Li, Juncheng and Tang, Siliang and Zhang, Hanwang and Zhuang, Yueting},
    year={2024},
    journal={arXiv preprint arXiv:2411.15738}
}

@inproceedings{Brooks2022InstructPix2PixLT,
    title={InstructPix2Pix: Learning to Follow Image Editing Instructions},
    author={Tim Brooks and Aleksander Holynski and Alexei A. Efros},
    year={2023},
    booktitle={CVPR},
}

@article{fu2023mme,
  title={MME: A Comprehensive Evaluation Benchmark for Multimodal Large Language Models},
  author={Fu, Chaoyou and Chen, Peixian and Shen, Yunhang and Qin, Yulei and Zhang, Mengdan and Lin, Xu and Yang, Jinrui and Zheng, Xiawu and Li, Ke and Sun, Xing and others},
  journal={arXiv preprint arXiv:2306.13394},
  year={2023}
}

@inproceedings{liu2024mmbench,
  title={Mmbench: Is your multi-modal model an all-around player?},
  author={Liu, Yuan and Duan, Haodong and Zhang, Yuanhan and Li, Bo and Zhang, Songyang and Zhao, Wangbo and Yuan, Yike and Wang, Jiaqi and He, Conghui and Liu, Ziwei and others},
  booktitle={ECCV},
  year={2024},
}

@inproceedings{yu2024mm,
  title={MM-Vet: Evaluating Large Multimodal Models for Integrated Capabilities},
  author={Yu, Weihao and Yang, Zhengyuan and Li, Linjie and Wang, Jianfeng and Lin, Kevin and Liu, Zicheng and Wang, Xinchao and Wang, Lijuan},
  booktitle={ICML},
  year={2024},
}

@inproceedings{yue2024mmmu,
  title={Mmmu: A massive multi-discipline multimodal understanding and reasoning benchmark for expert agi},
  author={Yue, Xiang and Ni, Yuansheng and Zhang, Kai and Zheng, Tianyu and Liu, Ruoqi and Zhang, Ge and Stevens, Samuel and Jiang, Dongfu and Ren, Weiming and Sun, Yuxuan and others},
  booktitle={CVPR},
  year={2024}
}

@inproceedings{lu23mathvista,
  title={MathVista: Evaluating Mathematical Reasoning of Foundation Models in Visual Contexts},
  author={Lu, Pan and Bansal, Hritik and Xia, Tony and Liu, Jiacheng and Li, Chunyuan and Hajishirzi, Hannaneh and Cheng, Hao and Chang, Kai-Wei and Galley, Michel and Gao, Jianfeng},
  year={2023},
  booktitle={NeurIPS Workshop on Mathematical Reasoning and AI}
}

@inproceedings{tong2024eyes,
  title={Eyes wide shut? exploring the visual shortcomings of multimodal llms},
  author={Tong, Shengbang and Liu, Zhuang and Zhai, Yuexiang and Ma, Yi and LeCun, Yann and Xie, Saining},
  booktitle={CVPR},
  pages={9568--9578},
  year={2024}
}

@inproceedings{ghosh2023geneval,
  title={Geneval: An object-focused framework for evaluating text-to-image alignment},
  author={Ghosh, Dhruba and Hajishirzi, Hannaneh and Schmidt, Ludwig},
  booktitle={NeurIPS},
  year={2023}
}

@article{molybog2023theory,
  title={A theory on adam instability in large-scale machine learning},
  author={Molybog, Igor and Albert, Peter and Chen, Moya and DeVito, Zachary and Esiobu, David and Goyal, Naman and Koura, Punit Singh and Narang, Sharan and Poulton, Andrew and Silva, Ruan and others},
  journal={arXiv preprint arXiv:2304.09871},
  year={2023}
}

@article{
    mot,
    title={Mixture-of-Transformers: A Sparse and Scalable Architecture for Multi-Modal Foundation Models},
    author={Weixin Liang and LILI YU and Liang Luo and Srini Iyer and Ning Dong and Chunting Zhou and Gargi Ghosh and Mike Lewis and Wen-tau Yih and Luke Zettlemoyer and Xi Victoria Lin},
    journal={TMLR},
    year={2025},
}

@article{huang2023t2i,
  title={T2i-compbench: A comprehensive benchmark for open-world compositional text-to-image generation},
  author={Huang, Kaiyi and Sun, Kaiyue and Xie, Enze and Li, Zhenguo and Liu, Xihui},
  journal={Advances in Neural Information Processing Systems},
  volume={36},
  pages={78723--78747},
  year={2023}
}

@article{deng2025emerging,
  title={Emerging properties in unified multimodal pretraining},
  author={Deng, Chaorui and Zhu, Deyao and Li, Kunchang and Gou, Chenhui and Li, Feng and Wang, Zeyu and Zhong, Shu and Yu, Weihao and Nie, Xiaonan and Song, Ziang and others},
  journal={arXiv preprint arXiv:2505.14683},
  year={2025}
}

@inproceedings{zhang2023blind,
  title={Blind image quality assessment via vision-language correspondence: A multitask learning perspective},
  author={Zhang, Weixia and Zhai, Guangtao and Wei, Ying and Yang, Xiaokang and Ma, Kede},
  booktitle={Proceedings of the IEEE/CVF conference on computer vision and pattern recognition},
  pages={14071--14081},
  year={2023}
}

@article{xu2023imagereward,
  title={Imagereward: Learning and evaluating human preferences for text-to-image generation},
  author={Xu, Jiazheng and Liu, Xiao and Wu, Yuchen and Tong, Yuxuan and Li, Qinkai and Ding, Ming and Tang, Jie and Dong, Yuxiao},
  journal={Advances in Neural Information Processing Systems},
  volume={36},
  pages={15903--15935},
  year={2023}
}

@inproceedings{lin2024evaluating,
  title={Evaluating text-to-visual generation with image-to-text generation},
  author={Lin, Zhiqiu and Pathak, Deepak and Li, Baiqi and Li, Jiayao and Xia, Xide and Neubig, Graham and Zhang, Pengchuan and Ramanan, Deva},
  booktitle={European Conference on Computer Vision},
  pages={366--384},
  year={2024},
  organization={Springer}
}

@article{guo2025can,
  title={Can We Generate Images with CoT? Let's Verify and Reinforce Image Generation Step by Step},
  author={Guo, Ziyu and Zhang, Renrui and Tong, Chengzhuo and Zhao, Zhizheng and Huang, Rui and Zhang, Haoquan and Zhang, Manyuan and Liu, Jiaming and Zhang, Shanghang and Gao, Peng and others},
  journal={arXiv preprint arXiv:2501.13926},
  year={2025}
}

@article{fang2025got,
  title={Got: Unleashing reasoning capability of multimodal large language model for visual generation and editing},
  author={Fang, Rongyao and Duan, Chengqi and Wang, Kun and Huang, Linjiang and Li, Hao and Yan, Shilin and Tian, Hao and Zeng, Xingyu and Zhao, Rui and Dai, Jifeng and others},
  journal={arXiv preprint arXiv:2503.10639},
  year={2025}
}

@article{duan2025got,
  title={Got-r1: Unleashing reasoning capability of mllm for visual generation with reinforcement learning},
  author={Duan, Chengqi and Fang, Rongyao and Wang, Yuqing and Wang, Kun and Huang, Linjiang and Zeng, Xingyu and Li, Hongsheng and Liu, Xihui},
  journal={arXiv preprint arXiv:2505.17022},
  year={2025}
}

@article{zhou2024calibrated,
  title={Calibrated self-rewarding vision language models},
  author={Zhou, Yiyang and Fan, Zhiyuan and Cheng, Dongjie and Yang, Sihan and Chen, Zhaorun and Cui, Chenhang and Wang, Xiyao and Li, Yun and Zhang, Linjun and Yao, Huaxiu},
  journal={Advances in Neural Information Processing Systems},
  volume={37},
  pages={51503--51531},
  year={2024}
}

@article{choi2024self,
  title={Self-improving robust preference optimization},
  author={Choi, Eugene and Ahmadian, Arash and Geist, Matthieu and Pietquin, Oilvier and Azar, Mohammad Gheshlaghi},
  journal={arXiv preprint arXiv:2406.01660},
  year={2024}
}

@article{chen2025blip3,
  title={Blip3-o: A family of fully open unified multimodal models-architecture, training and dataset},
  author={Chen, Jiuhai and Xu, Zhiyang and Pan, Xichen and Hu, Yushi and Qin, Can and Goldstein, Tom and Huang, Lifu and Zhou, Tianyi and Xie, Saining and Savarese, Silvio and others},
  journal={arXiv preprint arXiv:2505.09568},
  year={2025}
}

@article{sun2025t2i,
  title={T2I-ReasonBench: Benchmarking Reasoning-Informed Text-to-Image Generation},
  author={Sun, Kaiyue and Fang, Rongyao and Duan, Chengqi and Liu, Xian and Liu, Xihui},
  journal={arXiv preprint arXiv:2508.17472},
  year={2025}
}

@article{rafailov2023direct,
  title={Direct preference optimization: Your language model is secretly a reward model},
  author={Rafailov, Rafael and Sharma, Archit and Mitchell, Eric and Manning, Christopher D and Ermon, Stefano and Finn, Chelsea},
  journal={Advances in neural information processing systems},
  volume={36},
  pages={53728--53741},
  year={2023}
}

@misc{xie2025reconstructionalignmentimprovesunified,
      title={Reconstruction Alignment Improves Unified Multimodal Models}, 
      author={Ji Xie and Trevor Darrell and Luke Zettlemoyer and XuDong Wang},
      year={2025},
      eprint={2509.07295},
      archivePrefix={arXiv},
      primaryClass={cs.CV},
      url={https://arxiv.org/abs/2509.07295}, 
}

@article{wu2025qwen,
  title={Qwen-image technical report},
  author={Wu, Chenfei and Li, Jiahao and Zhou, Jingren and Lin, Junyang and Gao, Kaiyuan and Yan, Kun and Yin, Sheng-ming and Bai, Shuai and Xu, Xiao and Chen, Yilei and others},
  journal={arXiv preprint arXiv:2508.02324},
  year={2025}
}

@article{xie2025show,
  title={Show-o2: Improved Native Unified Multimodal Models},
  author={Xie, Jinheng and Yang, Zhenheng and Shou, Mike Zheng},
  journal={arXiv preprint arXiv:2506.15564},
  year={2025}
}

@article{xue2025dancegrpo,
  title={DanceGRPO: Unleashing GRPO on Visual Generation},
  author={Xue, Zeyue and Wu, Jie and Gao, Yu and Kong, Fangyuan and Zhu, Lingting and Chen, Mengzhao and Liu, Zhiheng and Liu, Wei and Guo, Qiushan and Huang, Weilin and others},
  journal={arXiv preprint arXiv:2505.07818},
  year={2025}
}

@article{wang2024diffusion,
  title={Diffusion feedback helps clip see better},
  author={Wang, Wenxuan and Sun, Quan and Zhang, Fan and Tang, Yepeng and Liu, Jing and Wang, Xinlong},
  journal={arXiv preprint arXiv:2407.20171},
  year={2024}
}

@article{wang2024reconstructive,
  title={Reconstructive visual instruction tuning},
  author={Wang, Haochen and Zheng, Anlin and Zhao, Yucheng and Wang, Tiancai and Ge, Zheng and Zhang, Xiangyu and Zhang, Zhaoxiang},
  journal={arXiv preprint arXiv:2410.09575},
  year={2024}
}

@article{hong2025reinforcing,
  title={Reinforcing Multimodal Understanding and Generation with Dual Self-rewards},
  author={Hong, Jixiang and Zhang, Yiran and Wang, Guanzhong and Liu, Yi and Wen, Ji-Rong and Yan, Rui},
  journal={arXiv preprint arXiv:2506.07963},
  year={2025}
}

@inproceedings{radford2021learning,
  title={Learning transferable visual models from natural language supervision},
  author={Radford, Alec and Kim, Jong Wook and Hallacy, Chris and Ramesh, Aditya and Goh, Gabriel and Agarwal, Sandhini and Sastry, Girish and Askell, Amanda and Mishkin, Pamela and Clark, Jack and others},
  booktitle={International conference on machine learning},
  pages={8748--8763},
  year={2021},
  organization={PmLR}
}

@poceedings{wang2023detecting,
  title={Detecting everything in the open world: Towards universal object detection},
  author={Wang, Zhenyu and Li, Yali and Chen, Xi and Lim, Ser-Nam and Torralba, Antonio and Zhao, Hengshuang and Wang, Shengjin},
  booktitle={Proceedings of the IEEE/CVF Conference on Computer Vision and Pattern Recognition},
  pages={11433--11443},
  year={2023}
}

@article{mao2025unirl,
  title={UniRL: Self-Improving Unified Multimodal Models via Supervised and Reinforcement Learning},
  author={Mao, Weijia and Yang, Zhenheng and Shou, Mike Zheng},
  journal={arXiv preprint arXiv:2505.23380},
  year={2025}
}

@inproceedings{kirillov2023segment,
  title={Segment anything},
  author={Kirillov, Alexander and Mintun, Eric and Ravi, Nikhila and Mao, Hanzi and Rolland, Chloe and Gustafson, Laura and Xiao, Tete and Whitehead, Spencer and Berg, Alexander C and Lo, Wan-Yen and others},
  booktitle={Proceedings of the IEEE/CVF international conference on computer vision},
  pages={4015--4026},
  year={2023}
}

@article{lin2025uniworld,
  title={Uniworld: High-resolution semantic encoders for unified visual understanding and generation},
  author={Lin, Bin and Li, Zongjian and Cheng, Xinhua and Niu, Yuwei and Ye, Yang and He, Xianyi and Yuan, Shenghai and Yu, Wangbo and Wang, Shaodong and Ge, Yunyang and others},
  journal={arXiv preprint arXiv:2506.03147},
  year={2025}
}

@article{han2025self,
  title={Self-Contradiction as Self-Improvement: Mitigating the Generation-Understanding Gap in MLLMs},
  author={Han, Yujin and Chen, Hao and Han, Andi and Wang, Zhiheng and Lin, Xinyu and Zhang, Yingya and Zhang, Shiwei and Zou, Difan},
  journal={arXiv preprint arXiv:2507.16663},
  year={2025}
}

@article{liao2025langbridge,
  title={LangBridge: Interpreting Image as a Combination of Language Embeddings},
  author={Liao, Jiaqi and Niu, Yuwei and Meng, Fanqing and Li, Hao and Tian, Changyao and Du, Yinuo and Xiong, Yuwen and Li, Dianqi and Zhu, Xizhou and Yuan, Li and others},
  journal={arXiv preprint arXiv:2503.19404},
  year={2025}
}

@inproceedings{qu2025silmm,
  title={Silmm: Self-improving large multimodal models for compositional text-to-image generation},
  author={Qu, Leigang and Li, Haochuan and Wang, Wenjie and Liu, Xiang and Li, Juncheng and Nie, Liqiang and Chua, Tat-Seng},
  booktitle={Proceedings of the Computer Vision and Pattern Recognition Conference},
  pages={18497--18508},
  year={2025}
}

\clearpage

\appendix
\section{Detail Settings}
\label{app:Detail}
Following the configuration of stage 4 from the \textbf{Bagel}~\cite{deng2025emerging} framework during our post-training phase, we employed the \textbf{AdamW} optimizer~\cite{loshchilov2017decoupled}, configured with momentum parameters $\beta_{1}=0.9$ and $\beta_{2}=0.95$. Drawing inspiration from~\cite{molybog2023theory}, we set the epsilon value to $1.0\times10^{-15}$ to mitigate loss spikes. When we increase the resolution during generation, we also adjust the diffusion timestep from 1.0 to 4.0, which helps maintain a stable noise-level distribution. We chose a constant learning rate, as this approach, as suggested by~\cite{hu2024minicpm}, simplifies the scaling of training data without needing to restart the training process. These empirical observations, along with established practices for large-scale model training~\cite{goyal2017accurate, hoffmann2022training, kaplan2020scaling,liao2025langbridge}, informed our final training protocol.

Our model architecture builds upon the standard Transformer~\cite{transformer} and ViT~\cite{vit} paradigms, incorporating modern enhancements for stability and efficiency, such as RMS Layer Normalization~\cite{zhang2019root}, GLU variants for activation functions~\cite{shazeer2020glu}, RoPE~\cite{su2024roformer}, and GQA~\cite{ainslie2023gqa}. The generative process is fundamentally based on principles from diffusion process~\cite{ddpm, sohl2015deep,song1}, and utilizes classifier-free guidance~\cite{ho2022classifier} within a latent space~\cite{rombach2022high} for high-resolution synthesis. The complete training recipe is summarized in \Cref{tab:training_recipe}.

\begin{table}[t] 
\centering
\caption{Training recipe of SRUM. We report the specific hyperparameter settings used during the post-training stage.}
\label{tab:training_recipe}
\small
\begin{tabular}{lc} 
\toprule
Hyperparameters & Post-training \\
\midrule
Learning rate & $2.5 \times 10^{-5}$ \\
LR scheduler & Constant \\
Weight decay & 0.0 \\
Gradient norm clip & 1.0 \\
Optimizer & AdamW ($\beta_1=0.9, \beta_2=0.95, \epsilon=10^{-15}$) \\
Warm-up steps & 500 \\
Max context window & 40k \\
Gen resolution (min short side, max long side) & (512, 1024) \\
Diffusion timestep shift & 4.0 \\
\bottomrule
\end{tabular}
\end{table}

In~\Cref{sec:data_gen}, we explain how detection boxes are generated. Bagel uses an external localization aid (SAM), while BLIP3o can rely on its native grounding capabilities. In both cases, the UMM verifies the boxes and supplies the semantic reward, so the external component is not an external reward model. The choice of localization source can be guided by the model's performance on grounding benchmarks such as RefCOCO.

\section{Definition and Calculation of Average Activation Strength}
\label{sec:appendix_activation}

To investigate the internal functional mechanisms of different training methods, we introduce the metric of \textit{Average Activation Strength}. This metric is designed to quantify the overall activity level of a predefined functional neural cluster when the model is performing a specific type of task. This appendix provides a detailed definition, mathematical formulation, and the statistical implementation procedure. The \textbf{Average Activation Strength} is defined as the mean activation value of all neurons within a specific functional cluster, averaged over an entire dataset for a given task. The calculation involves a two-level averaging process:
\begin{enumerate}
    \item \textbf{Intra-Cluster Average:} For a single input sample, we compute the mean of the activation values of all neurons belonging to the target cluster.
    \item \textbf{Dataset-Wide Average:} We then average these single-sample cluster means across all samples in the entire task dataset.
\end{enumerate}
This metric reflects the degree of engagement of a functional cluster (e.g., the ``Understanding Cluster'') while processing a certain category of tasks (e.g., ``Generation Tasks''). A higher value indicates that the cluster is more strongly and broadly activated for that task.

To formalize this definition, we first introduce the following notation:
\begin{itemize}
    \item $M$: A specific neural network model (e.g., Base, SFT, or SRUM).
    \item $C_k$: A functional neural cluster $k$ (e.g., $C_{\text{understand}}$ or $C_{\text{generate}}$), which is a set of specific neuron indices.
    \item $|C_k|$: The number of neurons in cluster $C_k$.
    \item $D_T$: The dataset for a specific task type $T$ (e.g., $D_{\text{understanding}}$ or $D_{\text{generation}}$).
    \item $|D_T|$: The number of samples in the dataset $D_T$.
    \item $x$: An individual input sample from the dataset, where $x \in D_T$.
    \item $a_i(x)$: The activation value of neuron $i$ in model $M$ given the input $x$, where $i \in C_k$. This typically refers to the output of a neuron after its activation function (e.g., ReLU or GeLU) has been applied.
\end{itemize}

For a single input sample $x$, the average activation strength of a cluster $C_k$, denoted as $S_{\text{sample}}$, is calculated as:
\begin{equation}
    S_{\text{sample}}(M, C_k, x) = \frac{1}{|C_k|} \sum_{i \in C_k} a_i(x)
\end{equation}

The final \textbf{Average Activation Strength} of cluster $C_k$ for model $M$ over the entire dataset $D_T$, denoted as $S_{\text{final}}$, is the expected value of $S_{\text{sample}}$ over all samples. In practice, this is estimated by averaging across the dataset:
\begin{equation}
    S_{\text{final}}(M, C_k, D_T) = \frac{1}{|D_T|} \sum_{x \in D_T} S_{\text{sample}}(M, C_k, x) = \frac{1}{|D_T||C_k|} \sum_{x \in D_T} \sum_{i \in C_k} a_i(x)
\end{equation}
This $S_{\text{final}}$ value corresponds to the height of each bar in the activation figures. Algorithm details can be seen in~\Cref{alg:aas_complete_single_threshold}.

\begin{algorithm*} 
\caption{Calculation of Average Activation Strength (with Single Threshold-Based Cluster Definition)}
\label{alg:aas_complete_single_threshold}
\begin{minipage}{0.48\textwidth} 
\begin{algorithmic}[1]
    \State \textbf{Step 1: Define Neuron Clusters}
    \State \quad \textbf{Require:}
    \State \quad \quad $M_{\text{base}}$: Base model.
    \State \quad \quad $D_{\text{und}}$: Understanding dataset.
    \State \quad \quad $D_{\text{gen}}$: Generation dataset.
    \State \quad \quad $\tau_{\text{act}}$: Activation percentile threshold (\%).
    \State \quad \textbf{Ensure:} $C_{\text{understand}}$, $C_{\text{generate}}$.
    \State
    \State \quad \textit{// (1.1) Collect mean activations}
    \State \quad Let $N$ be set of FFN neurons.
    \State \quad Init maps $\mu_{\text{und}}, \mu_{\text{gen}}, \mu_{\max}$.
    \For{each neuron $n \in N$}
        \State $\mu_{\text{und}}[n] \gets \text{mean}_{x \in D_{\text{und}}} a_n(x)$
        \State $\mu_{\text{gen}}[n] \gets \text{mean}_{x \in D_{\text{gen}}} a_n(x)$
    \EndFor
    \State
    \State \quad \textit{// (1.2) Calculate max activation and threshold}
    \For{each neuron $n \in N$}
        \State $\mu_{\max}[n] \gets \max(\mu_{\text{und}}[n], \mu_{\text{gen}}[n])$
    \EndFor
    \State $V_{\text{act}} \gets \text{Percentile}(\{\mu_{\max}[n] \mid n \in N\}, \tau_{\text{act}})$
    \State
    \State \quad \textit{// (1.3) Filter clusters based on activation threshold and max activation task}
    \State \quad $C_{\text{understand}} \gets \emptyset$, $C_{\text{generate}} \gets \emptyset$
    \For{each neuron $n \in N$}
        \If{$\mu_{\max}[n] \ge V_{\text{act}}$} \Comment{Must be an active neuron}
            \If{$\mu_{\text{und}}[n] > \mu_{\text{gen}}[n]$} \Comment{More active for understanding}
                \State $C_{\text{und}} \leftarrow C_{\text{und}} \cup \{n\}$
            \ElsIf{$\mu_{\text{gen}}[n] > \mu_{\text{und}}[n]$} \Comment{More active for generation}
                \State $C_{\text{gen}} \leftarrow C_{\text{gen}} \cup \{n\}$
            \EndIf
        \EndIf
    \EndFor
    \State \quad \textit{// Clusters fixed}
\end{algorithmic}
\end{minipage}
\hfill 
\begin{minipage}{0.48\textwidth} 
\begin{algorithmic}[1] 
    \State \textbf{Step 2: Prepare Eval Data \& Model}
    \State \quad Prepare dataset (e.g., $D_{\text{und}}$).
    \State \quad Load model $M$.
    \State
    \State \textbf{Step 3: Forward Pass \& Log}
    \State \quad Init lists: und\_activ = [], gen\_activ = []
    \For{each sample $x \in D_{\text{understanding}}$}
        \State Forward pass $M(x)$.
        \State Record $a_i(x)$ for $i \in C_{\text{und}}, C_{\text{gen}}$.
        \State Calc sample avg activation:
        \State \quad $S_{\text{samp, und}} \gets \text{mean}_{i \in C_{\text{und}}} a_i(x)$
        \State \quad $S_{\text{samp, gen}} \gets \text{mean}_{i \in C_{\text{gen}}} a_i(x)$
        \State Append $S_{\text{samp, und}}$ to und\_activ
        \State Append $S_{\text{samp, gen}}$ to gen\_activ
    \EndFor
    \State
    \State \textbf{Step 4: Final Aggregation}
    \State \quad $S_{\text{final, und}} \gets \text{mean}(\text{und\_activ})$
    \State \quad $S_{\text{final, gen}} \gets \text{mean}(\text{gen\_activ})$
    \State \quad Output $S_{\text{final, und}}, S_{\text{final, gen}}$.
    \State
    \State \textbf{Step 5: Repeat Process}
    \State \quad Repeat Steps 3-4 using $D_{\text{gen}}$.
    \State \quad Repeat Steps 2-5 for each model.
\end{algorithmic}
\end{minipage}
\end{algorithm*}

\section{Data Curation}
\label{app:data_curation}
We leverage the training instructions from T2I-CompBench~\cite{huang2023t2i} to guide our image generation process. Specifically, we utilize the generation capabilities of UMMs~\cite{vilau2024, janus2024, show-o, dongdreamllm}, which are representative of the state-of-the-art in text-to-image synthesis~\cite{dalle3, imagen, SD3, flux, wu2025qwen}, to synthesize corresponding images based on these instructions. Subsequently, the understanding side of UMMs, which possesses powerful vision-language comprehension abilities akin to models like LLaVA, InternVL, and Gemini~\cite{llava-plus, internvl2, qwen2vl, gemini}, is employed to evaluate and score the generated images.

The capabilities of these models are built upon massive web-scale datasets~\cite{schuhmann2022laion, li2024omnicorpus} and canonical vision datasets~\cite{lin2014microsoft}, which are often enhanced with high-quality captioning and instruction-following data~\cite{sharma2018cc, li2024datacomprecap, liu2023visual}. Our prompting strategy for eliciting rewards is inspired by the methodologies used in instruction-based image editing~\cite{Brooks2022InstructPix2PixLT, wei2024omniedit, zhang2023magicbrush, yu2024anyedit, hui2024hqedit, bai2024humanedit}. The detailed data used in this evaluation are as follows:

\begin{table*}[!ht]
\centering
\begin{minipage}{1\textwidth}\vspace{0mm}
    \centering
    \begin{tcolorbox}
        \centering
        \hspace{-6mm}
        \begin{tabular}{p{1\textwidth}}
        \hspace{1mm}
        \begin{minipage}{1\textwidth}
        \textbf{Generated Prompt Content:}\\
        \begin{verbatim}
# TASK: Global Layout and Composition Analysis
You are an expert image analyst. 

Your task is to score the overall composition 
of an image based on a user's prompt. Focus solely 
on how the arrangement of elements and scene structure 
align with the prompt's spatial intent.

**Original Prompt:** "{original_prompt}"
---
## YOUR TASK & OUTPUT FORMAT
Provide a single score from **-1.0 to 1.0** and a brief reason.

* **Scoring Guide:**
* **1.0:** Perfect alignment with the prompt's 
spatial intent.
* **0.5 to 0.9:** Mostly correct layout 
with minor flaws.
* **-0.4 to 0.4:** Neutral. No specific spatial 
info in prompt, or generic layout.
* **-0.9 to -0.5:** Incorrect layout or 
contradictory to the prompt.
* **-1.0:** Fundamentally contradicts the 
prompt's spatial intent.

* **Output Lines:**
    `Score: [A single number between -1.0 and 1.0]'
    `Reason: [Your justification]'
---
## DIVERSE EXAMPLES

### Example 1 (Perfect Alignment)
Score: 0.95
Reason: The wide shot of a sunset over the ocean perfectly 
matches the prompt's implied composition.

### Example 2 (Contradictory Layout)
Score: -0.7
Reason: The cat is on the right of the dog, but the prompt 
asked for the cat on the left.
---
Begin your analysis now.
        \end{verbatim}
        \end{minipage}
        \end{tabular}
    \end{tcolorbox}
    \vspace{-2mm}
    \caption{\textbf{Documentation for \texttt{create\_global\_layout\_reward\_prompt}.}}
    \label{tab:func_global_layout_original}
    \end{minipage}
    \vspace{-2mm}
\end{table*}

\begin{table*}[!ht]
\centering
\begin{minipage}{1\textwidth}\vspace{0mm}
\small
    \centering
    \begin{tcolorbox}
        \centering
        \hspace{-6mm}
        \begin{tabular}{p{1\textwidth}}
        \hspace{1mm}
        \begin{minipage}{0.8\textwidth}
        \textbf{Generated Prompt Content:}\\
        \begin{verbatim}
# TASK: Integrated Region Analysis and Scoring
You are an expert AI image analyst. 
Your task is to analyze unlabeled regions in an image 
based on a user's prompt. 
For each region, you will perform a two-stage analysis.
**Original Prompt:** "{original_prompt}"
---
**UNLABELED REGIONS FOR YOUR ANALYSIS:**
{regions_text}
---
## YOUR TWO-STAGE TASK & OUTPUT FORMAT
For **every Region ID** listed above, 
you must perform the following steps.

### STAGE 1: Identify Object
First, identify the primary object within the bounding box.
* **Output Line:** 
`Identified Object: [Your description of the object]'

### STAGE 2: Score and Justify
Provide a single, overall score 
from **-1.0 to 1.0** that considers BOTH the object's 
**relevance** to the prompt and its **visual quality**. 
You must provide a clear reason for your score.

* **Scoring Guide:**
    * **1.0:** Perfect. The object is exactly what the...
    * **0.5 to 0.9:** Very good. A highly relevant object...
    * **-0.4 to 0.4:** Neutral/Acceptable. A moderately 
    relevant object, an object with mixed qualities...
    * **-0.9 to -0.5:** Bad. The object is irrelevant 
    and distracting, or it is a relevant object with 
    severe visual artifacts/flaws.
    * **-1.0:** Very Bad. The object actively 
    undermines...

* **Output Lines:**
    `Score: [A single number between -1.0 and 1.0]'
---
## EXAMPLE OUTPUT STRUCTURE
**Region ID: 1**
Identified Object: A running golden retriever.
Score: 0.95
---
Begin your analysis now.
        \end{verbatim}
        \end{minipage}
        \end{tabular}
    \end{tcolorbox}
    \vspace{-2mm}
    \caption{\textbf{Documentation for \texttt{create\_hybrid\_evaluation\_prompt}.}}
    \label{tab:func_hybrid_eval_original}
    \end{minipage}
    \vspace{-2mm}
\end{table*}

\begin{table}[htbp]
\centering
\caption{VLM Rewards for Prompt: ``a microwave hidden by a horse''}
\label{tab:vlm_rewards_horse}

\begin{tabular}{p{4.5cm} l c p{3.5cm}}
\toprule
\textbf{Object} & \textbf{Bounding Box (bbox)} & \textbf{Score} & \textbf{Reason} \\
\midrule

global\_layout\_reward & [0, 0, 1024, 1024] & 1.00 & The image perfectly aligns with the prompt's spatial intent by depicting a horse positioned in front of a microwave, effectively hiding it from view. The composition is well-executed, with the horse's body and legs obscuring the microwave, and the plain background ensuring focus on the interaction between the two elements. \\
\midrule 
A brown horse with a white blaze and white socks. & [164, 97, 957, 990] & 0.95 & -- \\
A brown horse with a white blaze and white socks. & [0, 0, 1023, 831] & 0.95 & -- \\
A brown horse with a white blaze and white socks. & [349, 28, 920, 880] & 0.95 & -- \\
A microwave. & [349, 28, 920, 389] & 0.50 & -- \\
The floor. & [0, 681, 1023, 1023] & 0.00 & -- \\
The floor. & [0, 838, 1023, 1023] & 0.00 & -- \\
A brown horse with a white blaze and white socks. & [422, 94, 748, 292] & 0.95 & -- \\
A brown horse with a white blaze and white socks. & [429, 589, 856, 795] & 0.95 & -- \\
A brown horse with a white blaze and white socks. & [430, 121, 848, 793] & 0.95 & -- \\
A brown horse with a white blaze and white socks. & [430, 607, 755, 780] & 0.95 & -- \\
\bottomrule
\end{tabular}
\end{table}

\newpage

\section{Failure Cases Study}
\label{fail}
We conducted an analysis of three failure cases:

\begin{enumerate}
    \item \textbf{The language model is unable to arrive at the correct answer.}
    Our prompt was:
    \begin{quote}
        ``Given the following mapping: 1 -- apple, 2 -- banana, 3 -- watermelon. Compute: $1 + 3 - 2 + 1$, then return the fruit corresponding to the result.''
    \end{quote}
    In this scenario, most language models answer incorrectly. Therefore, the generation module in this case can only generate ``apple.''

    \item \textbf{Causal multi-image generation.}
    Because the training data for Bagel rarely contains data representing causality in a single image, we are unable to achieve good results for this type of task. Our example was:
    \begin{quote}
        ``Generate a comparison image of British cities before and after the Industrial Revolution.''
    \end{quote}

    \item \textbf{Aesthetic generation issues.}
    Our method focuses on problems related to reasoning, knowledge, and composition. Consequently, aesthetics are not a primary consideration, which is also a common issue in existing models. Our example was:
    \begin{quote}
        ``Generate a particularly beautiful chair.''
    \end{quote}
\end{enumerate}

The top row shows our failure cases, and the bottom row shows the failure cases of nano-banana (current frontier model), illustrating that this failure is a systemic problem in generative models.
\begin{figure}[H]      
   \vspace{-1em}
  \hspace*{0.01\textwidth}
  \includegraphics[width=0.98\textwidth]{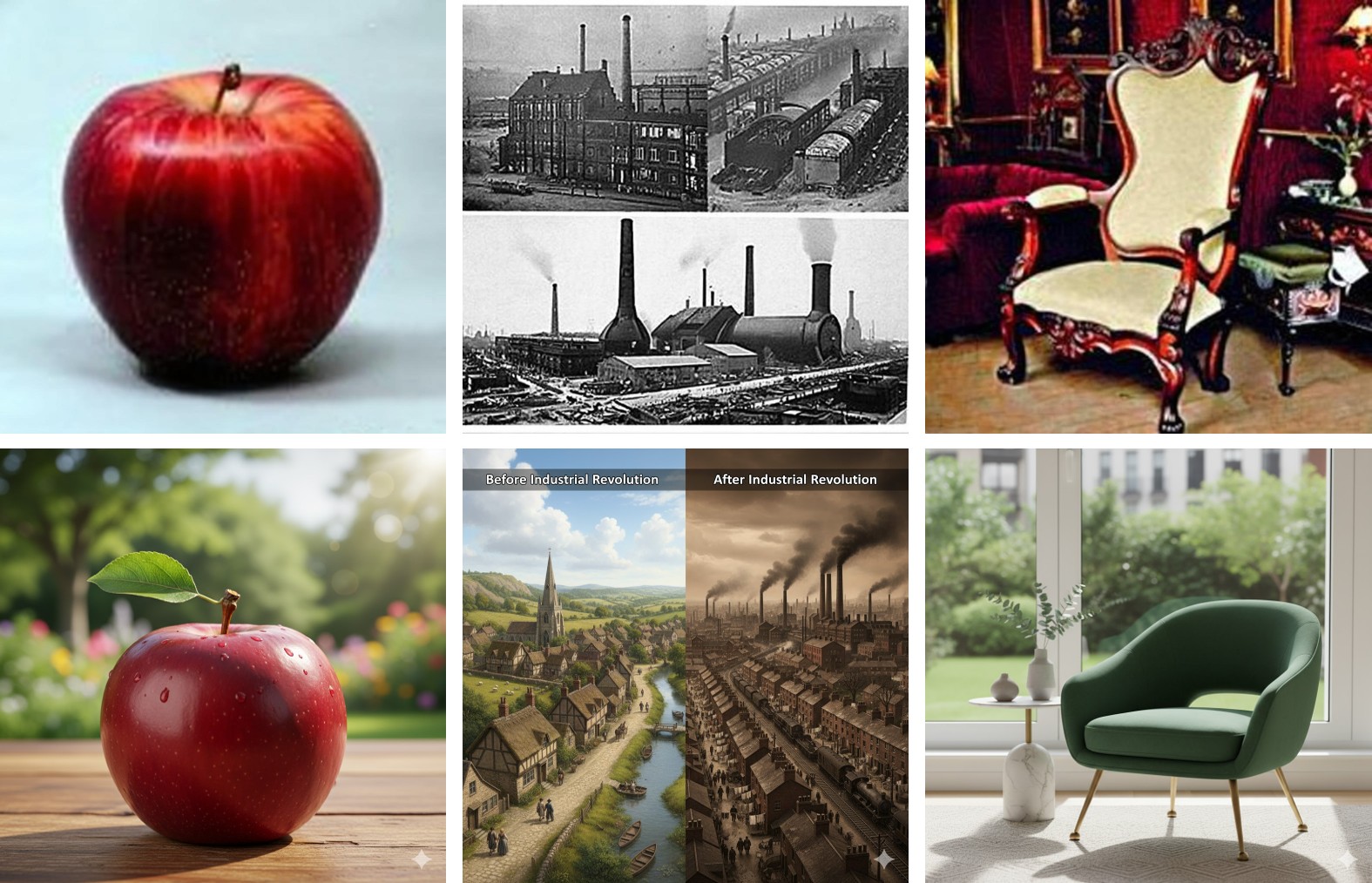}
   \vspace{-0.5em}
  \label{fig:fail}
\end{figure}
\end{document}